%% file: iclr2027_conference.tex
\documentclass{article} 
\usepackage{iclr2027_conference,times}

\input{math_commands.tex}

\usepackage{hyperref}
\usepackage{url}
\usepackage{booktabs}
\usepackage{graphicx}
\usepackage{amsmath,amssymb}
\usepackage{wrapfig}
\usepackage[nameinlink,capitalize]{cleveref}

\title{Reason in Style: Discovering and \\ Controlling Style in Language Models}

\author{Ioana Marinescu\thanks{Correspondence: \texttt{im2178@nyu.edu}} \\
NYU
\And
Eric Karl Oermann \\
NYU \& NYU Langone Health
\And
Kyunghyun Cho \\
NYU
}

\iclrfinalcopy 
\begin{document}

\maketitle

\begin{abstract}
Language models learn content and style jointly, making stylistic variation in their outputs difficult to identify and control. We study whether recurring styles in model responses can be discovered without supervision and explicitly controlled. We design an algorithm that learns to separate representations of content and style from language models' outputs and validate its effectiveness on math questions in a controlled setting. By applying this method to over 100K verified traces from nine distinct teacher models, we discover six recurring yet imbalanced styles.
We then fine-tune smaller student models to follow these styles when explicitly conditioned on them, using importance weighting to balance the contribution of the styles represented in the corpus.
This approach improves Pass@$k$ over standard fine-tuning on the same data across six math reasoning benchmarks, demonstrating that we can diversify the style of answers effectively.
We confirm that this also results in strong correspondence between requested and realized styles. We find that style affects correctness: the probability of solving a problem depends on the style we condition on, and different problems benefit from different styles.
In summary, our results show that stylistic variation in model-generated data can be discovered in an unsupervised way, and made explicit, providing a source of both control and improved reasoning performance.
\end{abstract}

\section{Introduction}

Language models are trained to predict the next token in a sequence, an
objective that does not explicitly distinguish between different aspects of
the text being modeled. In particular, it does not separate \emph{what} a
response says from \emph{how} it is expressed: semantic content is learned
jointly with stylistic properties such as organization, verbosity, formatting,
and lexical style. This entanglement becomes important when we want to analyze
or control model behavior. A generated response may contain the desired
semantic content while also exhibiting stylistic properties inherited from
pretraining or post-training data. Separating these factors would allow us to
characterize stylistic variation independently of content and control it
explicitly during training and generation.

We study this problem in mathematical reasoning, where the distinction between
content and style is particularly concrete. The same mathematical problem can
be solved correctly while the solution is expressed in different
ways. Different language models naturally produce such variation. However,
there is no predefined taxonomy specifying which styles are present in a
collection of model-generated solutions, and style is not necessarily tied to
model identity: different models may produce similar styles, while the same
model may produce multiple styles. Our goal is therefore to discover this
structure automatically and represent style separately from problem content.

The need to make this stylistic structure explicit becomes particularly important as language models are increasingly post-trained on model-generated data that provides complete reasoning traces~\citep{hsieh2023distilling,mukherjee2023orca,feng2024kpod}. Fine-tuned models can inherit the structural and lexical properties of their training traces~\citep{lippmann2025stylesubstance}, even though the training objective does not explicitly represent style. Instead, these properties are learned implicitly from the examples themselves, so styles that occur frequently in the training data can dominate the model's behavior. At the same time, stylistic mismatch between the training data and the model's prior can make fine-tuning less effective or even degrade performance~\citep{huang2026finetune}. Discovering and representing style explicitly would allow us to control which stylistic behaviors are transferred during fine-tuning and how strongly they influence the resulting model, rather than permit the model to implicitly inherit them from the data composition.

To isolate style in reasoning traces, we first learn a representation that separates stylistic variation from
problem-specific content using an autoencoder with an objective designed for
disentanglement. We validate this representation in a controlled setting in
which problems are answered in four known writing
styles. Although the style labels are never used during training, the learned
style representation recovers the hidden styles while the content
representation becomes invariant to them. We then apply the same
method to a dataset of 111{,}834 verified mathematical reasoning traces generated by nine
models from~\citet{hu2026scas}. Without assuming a predefined style taxonomy,
we discover six recurring styles expressed by different models for multiple questions.

\begin{figure}[!t]
    \centering
    \includegraphics[width=\textwidth]{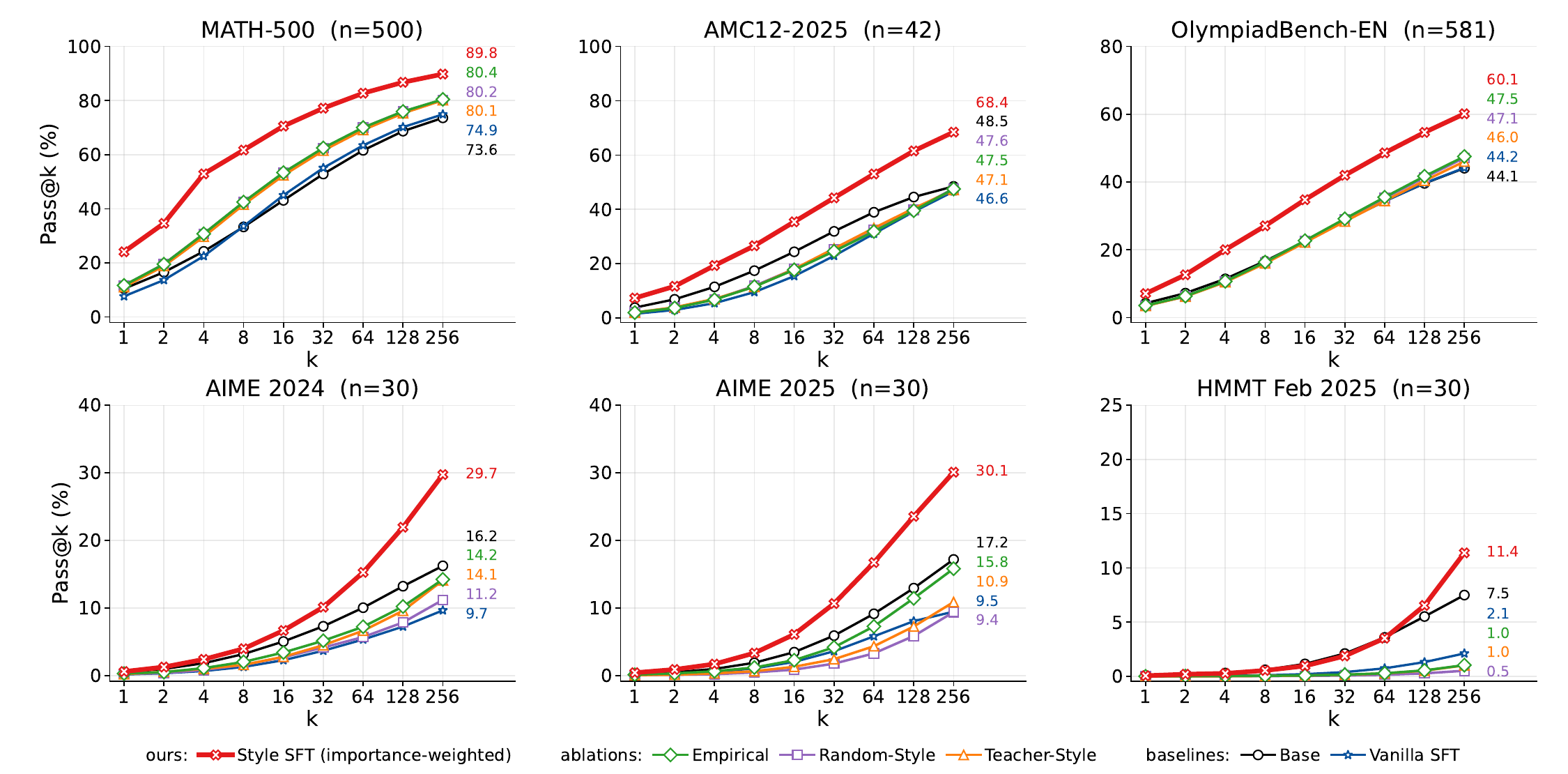}
    \caption{
Pass@$k$ for the Qwen3 0.6B student.
\textsc{Style SFT (Importance-Weighted)} is our full method, combining discovered style conditioning with importance-weighted SFT.
\textsc{Style SFT (Empirical)} conditions on the discovered style labels while retaining the empirical training distribution.
\textsc{Random-Style SFT} uses randomly assigned style labels as a control.
\textsc{Vanilla SFT} uses the same training traces without style information.
Our full method substantially improves Pass@$k$, while vanilla SFT often provides little benefit or degrades performance.
}
    \label{fig:06bpassk}
\end{figure}

We next use these discovered styles to make these models style-aware.
We label each training trace with its discovered style, and then fine-tune
Qwen3 students of three sizes using style prefixes and
importance weighting to balance the contribution of the styles observed
for each problem. The resulting students are controllable and exhibit meaningful use of the
discovered styles. Moreover, style-aware fine-tuning
substantially improves Pass@$k$ over standard fine-tuning on the same dataset across six mathematical
reasoning benchmarks (Figure~\ref{fig:06bpassk}). The correctness of a solution appears to
depend jointly on problem content and the style in which the model is
conditioned to reason.

Taken together, our results provide a first demonstration that stylistic
variation in model-generated reasoning can be discovered without supervision,
separated from problem content, and exposed as an explicit control variable
during fine-tuning and generation. Our approach shifts the role of style in language models from an incidental consequence of the training objective to a representable and controllable component of model behavior.

\section{Related Work}

\paragraph{Disentangling and controlling text style.}
A large body of work on text style transfer aims to modify how text is
expressed while preserving its semantic content. Many approaches explicitly
separate representations of content and style, including methods based on
adversarial objectives, information-theoretic constraints, and structured
latent spaces~\citep{shen2017style,john2019disentangled,cheng2020information,
sha2024closedloop,zhu2024styleflow,han2024disentangled}. However, these methods
typically assume a known style attribute, predefined domains, or a fixed set
of style categories. Our setting instead assumes no style labels or predefined
taxonomy. We use multiple independently generated solutions to the same
problem as weak supervision for shared content, allowing recurring variation
across solutions to be represented separately as style.

\paragraph{Style in model-generated training data.}
Model-generated reasoning traces exhibit systematic style variation in how solutions are organized and expressed across models and scales~\citep{chen2025thoughts}.
Trace-level analyses further identify recurring behaviors such as exploration,
verification, and backtracking~\citep{bachmann2026potential,
zhang2026overthinking}. \citet{lippmann2025stylesubstance} show that structural and lexical properties of teacher-generated reasoning traces are transferred to students. \citet{huang2026finetune} further show that stylistic mismatch between generated training data and the student model can affect fine-tuning
performance. These results suggest that the style of training data can shape the behavior of the resulting model. Our work makes the stylistic variation explicit by discovering the styles present in the data. 

\paragraph{Controlling solution modes.}
Sampling multiple solution trajectories can improve reasoning
performance~\citep{wang2023selfconsistency}, motivating methods that explicitly
encourage or control variation across reasoning paths.
\citet{wu2025modeconditioning} condition generation on distinct reasoning
modes using mode-specific prefixes and show that allocating samples across
modes can improve Pass@$k$. They further introduce gradient clustering to
discover modes without predefined labels; on multi-teacher data, these
clusters closely recover teacher identity.
Relatedly, \citet{jia2026training} learn global forking tokens that specialize
to different reasoning paths, allowing multiple solution modes to be preserved
and elicited during generation.
In our setting, teacher identity is not a reliable proxy for style:
different teachers can produce similar styles, while a single teacher can
produce multiple styles. We therefore learn stylistic structure directly from
the traces, and use the discovered styles as controllable variables during fine-tuning and generation.

\section{Unsupervised Learning of Style Representations}
\label{sec:method-ae}

Our goal is to separate information about the underlying problem from variation in how a solution is expressed. We exploit the fact that independently generated traces (answers) for the same question share problem-specific information but may differ in organization, notation, verbosity, verification, and other stylistic properties. We use an autoencoder with a disentanglement objective to learn separate style and content representations, inspired by~\citet{makino2025supervisedcontrastiveblockdisentanglement}.

\paragraph{Architecture.}
An encoder $E$ maps each trace $s$ to a representation $h(s)$, which is projected by two MLP heads into a content representation $c(s)$ and a style representation $z(s)$:
$
c(s)\in\mathbb{R}^{d_c},
z(s)\in\mathbb{R}^{d_z},
$
with $d_c=d_z=256$. 
A decoder reconstructs the original trace from both representations. We concatenate $[c(s);z(s)]$ and project it into $m=8$ continuous prefix tokens that condition an autoregressive decoder. Reconstruction requires the two representations jointly to retain sufficient information about the trace.

\paragraph{Training objective.}
We train the model with the following objective:
\begin{equation}
\label{eq:total-loss}
\mathcal{L}
=
\lambda_c\mathcal{L}_c
+
\lambda_{\mathrm{orth}}\mathcal{L}_{\mathrm{orth}}
+
\lambda_{\mathrm{rec}}\mathcal{L}_{\mathrm{rec}}
+
\lambda_{\mathrm{var}}\mathcal{L}_{\mathrm{var}}
+
\lambda_{\mathrm{cov}}\mathcal{L}_{\mathrm{cov}}
+
\lambda_{\mathrm{div}}\mathcal{L}_{\mathrm{div}}.
\end{equation}


\paragraph{Content agreement ($\mathcal{L}_c$).}
We use an InfoNCE objective~\citep{oord2018cpc,chen2020simclr} in which the two traces generated for the same question form a positive pair and traces from other questions in the minibatch serve as negatives. This encourages $c$ to capture information that is shared across independently generated solutions to the same problem.

\paragraph{Content--style decorrelation ($\mathcal{L}_{\mathrm{orth}}$).}
We penalize the squared cosine similarity between $c(s)$ and $z(s)$ for the same trace. This discourages the two heads from learning aligned representations and encourages $z$ to capture information complementary to the question-specific content represented in $c$.

\paragraph{Reconstruction ($\mathcal{L}_{\mathrm{rec}}$).}
The decoder reconstructs the original trace from the continuous prefix derived from $[c(s);z(s)]$. This ensures that the two representations remain jointly informative about the observed solution rather than discarding trace-level information.

\paragraph{Variance and covariance regularization ($\mathcal{L}_{\mathrm{var}},\mathcal{L}_{\mathrm{cov}}$).}
To prevent the style representation from collapsing, we use VICReg-style variance and covariance regularization~\citep{bardes2022vicreg}. The variance term encourages each dimension of $z$ to maintain nontrivial variation across the minibatch, while the covariance term discourages redundant dimensions from becoming highly correlated.

\paragraph{Within-question diversity ($\mathcal{L}_{\mathrm{div}}$).}
Finally, we discourage the style representations of the two solutions to the same question from collapsing together. While $\mathcal{L}_c$ pulls same-question content representations together, $\mathcal{L}_{\mathrm{div}}$ penalizes positive cosine similarity between their corresponding $z$ representations. This prevents $z$ from simply becoming another representation of question identity and leaves room for differences between independently generated solutions to be represented as style.

Together, the losses encourage $c$ to capture information that is stable across solutions to the same problem, while $z$ captures complementary trace-level variation. Full definitions of each objective, loss weights, and implementation details are provided in Appendix~\ref{app:style-representation-details}.

\section{Style Discovery}

\subsection{Recovering Known Styles}
\label{sec:known-styles}

Before applying our representation to unlabeled teacher traces, we first test whether it can recover a known source of stylistic variation without using style labels during training.

\paragraph{Controlled setup.}
We prompt Gemini 3.1 Pro~\citep{gemini31pro2026} to solve 300 math problems in four explicitly specified writing styles: academic prose, tutor-like explanations, algorithmic step-by-step lists, and equation-dominated solutions. Each problem is generated in all four styles, yielding 1{,}200 traces and ensuring that style is not confounded with question identity. Style labels are used only for evaluation. We train the autoencoder using the objective described above and obtain a content representation $c$ and a style representation $z$.

\paragraph{Recovering style without supervision.}
The learned representation separates stylistic variation from problem content. Clustering $z$ with $k$-means using $k=4$ recovers the four ground-truth styles with ARI $0.77$ and NMI $0.78$. In contrast, clustering $c$ shows essentially no agreement with style. A linear probe similarly predicts style from $z$ with $96.7\%$ accuracy, while accuracy from $c$ is $23.3\%$, near the $25\%$ chance level. This separation does not emerge from reconstruction alone. With reconstruction-only training, style is encoded in both representations: style ARI is $0.85$ from $z$ and $0.91$ from $c$. The frozen pretrained encoder also contains stylistic information, but its representation is poorly organized for unsupervised recovery, with style ARI $0.29$. Thus, our objective does not simply preserve style information already present in the encoder; it organizes this information in $z$ while removing it from $c$. We show additional analysis of the learned embeddings in Appendix~\ref{app:known-styles}.
These results validate $z$ as a representation in which known stylistic variation can be recovered without style supervision. We next apply the same procedure to unlabeled traces from the multi-teacher reasoning corpus.

\subsection{Styles in Multi-Teacher Reasoning Data}
\label{sec:scas-styles}

We use the SCAS dataset from \citet{hu2026scas}, which contains over 100K verified mathematical reasoning traces generated by nine teacher models. Details on the dataset and autoencoder training are shown in Appendix~\ref{app:teacher-styles}.
We encode each teacher trace using the learned style representation $z(s)$ and fit a Gaussian mixture model for clustering. We use $K=6$ clusters, chosen on the validation set, in our main experiments. We order the clusters by mean trace length and denote them $s_1,\ldots,s_6$.

\paragraph{What styles emerge?}
\label{sec:styles-emerge}

The discovered styles range from short textbook-like solutions ($s_1$) to long think-block traces ($s_6$), with intermediate styles differing in formatting, use of markdown, verification, backtracking, and other stylistic features. Importantly, the styles are imbalanced in the generated training data. The dominant style, $s_5$, accounts for $41.3\%$ of all traces, whereas each of the remaining styles accounts for only $9.0$--$15.3\%$. Thus, although the teacher pool contains multiple ways to express solutions, standard fine-tuning exposes the student to them at different frequencies. Full teacher--style distributions are reported in Appendix~\ref{app:teacher-style-composition}.

\begin{wraptable}{r}{0.54\columnwidth}
\vspace{-2.1\baselineskip}
\centering
\caption{
Teacher composition of each discovered style (\%; rows sum to 100).
}
\label{tab:style-composition-full}
\scriptsize
\setlength{\tabcolsep}{2.2pt}
\resizebox{\linewidth}{!}{%
\begin{tabular}{lccccccccc}
\toprule
Style & gemma & gpt-5 & gpt-oss & llama & olmo & phi-4 & qwen2.5 & qwen3 & qwen3.5 \\
\midrule
$s_1$ & 30.2 & 0.3 & 0.1 & 34.5 & 4.0 & 0.0 & 28.6 & 0.4 & 1.9 \\
$s_2$ & 16.4 & 10.8 & 5.5 & 16.0 & 12.4 & 8.1 & 13.2 & 6.5 & 11.0 \\
$s_3$ & 0.8 & 29.6 & 34.7 & 0.0 & 18.8 & 9.3 & 2.6 & 0.3 & 3.9 \\
$s_4$ & 2.4 & 24.0 & 23.3 & 0.1 & 19.2 & 18.7 & 5.2 & 2.6 & 4.5 \\
$s_5$ & 10.0 & 10.0 & 11.9 & 8.7 & 12.0 & 7.9 & 10.9 & 16.6 & 12.1 \\
$s_6$ & 1.9 & 4.9 & 0.7 & 3.5 & 4.0 & 31.8 & 0.1 & 24.2 & 28.8 \\
\bottomrule
\end{tabular}%
}
\vspace{-2\baselineskip}
\end{wraptable}

Styles appear across multiple teachers. For example, llama, gemma, and qwen2.5
frequently produce $s_1$, whereas phi-4, qwen3, and qwen3.5 mostly $s_6$. The normalized mutual information between teacher identity and style is only $0.17$. In particular, the dominant $s_5$ cluster contains substantial mass from all nine teachers. 

Moreover, the styles do not identify the underlying mathematical problem. Because each question has nine independently generated teacher traces, we can measure stylistic variation while holding the problem fixed. Only $0.85\%$ of questions have all nine solutions assigned to the same style. A question contains $3.38$ distinct styles on average, and $83\%$ of questions contain at least three. Agreement between question identity and style is correspondingly low (AMI $=0.049$, ARI $\approx 0$). Examples of traces for each style are shown in Appendix~\ref{app:teacher-style-examples}. 

\paragraph{Comparison with LLM-based style discovery.}
As a simple alternative to learning a style representation, we also ask a
language model to discover recurring styles directly from the traces. Because
direct prompting is limited by context length, we apply this baseline to a
balanced subset of 180 traces, with 20 traces from each of the nine models.
The resulting partition is strongly aligned with model identity
(ARI $=0.70$), whereas our discovered styles have much weaker agreement with
model identity (ARI $=0.06$). Thus, direct LLM-based discovery primarily
recovers model-specific differences on this subset, rather than the
cross-model stylistic structure captured by our autoencoder representation. Details for this experiment are shown in Appendix~\ref{app:llmstyledicovery}.

\section{Style-aware fine-tuning}
\label{sec:style-aware-distillation}

We next use the discovered styles to make models explicitly aware of style. We use the same dataset for fine-tuning. We first pass every training trace through the learned autoencoder, extract its style representation $z$ and identify its corresponding style cluster, then assign the resulting style label to the trace. We then prepend this label to the trace during supervised fine-tuning (SFT), allowing the student to condition its generation on a requested style at inference time.

Simply adding style labels, however, leaves the empirical style distribution unchanged. Because the discovered styles are imbalanced, the student still receives more supervision for common styles than for rare ones. In practice, this \textsc{Style SFT (Empirical)} baseline performs similarly to vanilla SFT, suggesting that exposing style labels alone is not sufficient to make effective use of the different modes present in the training data.

To mitigate this, we combine style conditioning with importance weighting: we reweigh the contribution of training traces to the training objective evenly among styles. This gives underrepresented styles greater influence during training while preserving all available reasoning traces. At inference, we generate under each style prefix, exposing multiple style-conditioned solution distributions.

\subsection{Importance-Weighted Style-Aware Training}
\label{sec:style-aware-training}

The discovered styles are unevenly represented not only globally, but also among the teacher traces available for an individual question. Let $S_i$ denote the set of styles observed among the teacher traces for question $x_i$. We define a uniform target distribution over the styles observed for $x_i$,
$
p(c\mid x_i)
=
{\mathbb{I}(c\in S_i)}/
{|S_i|},
$
and assign each trace an importance weight
$
w(c\mid x_i)
\propto
{p(c\mid x_i)}/
{q(c\mid x_i)},
$
where $q(c\mid x_i)$ is a smoothed estimate of the empirical style distribution for question $x_i$. Thus, styles that are overrepresented for a question receive smaller weights, while underrepresented styles receive larger weights. We retain every training trace and only change its contribution to the negative log-likelihood objective. We mean-normalize the weights to one within the training set. Full details of the proposal distribution, smoothing, and weighting objective are given in Appendix~\ref{app:importance-weighting}.

\paragraph{Training setup.}
We fine-tune Qwen3-0.6B-Base, Qwen3-1.7B-Base, and Qwen3-4B-Base students on the same multi-teacher corpus using full-parameter SFT. All models are trained with a maximum sequence length of $4096$ and an effective batch size of $256$ for 40 epochs with a constant learning rate of $10^{-5}$. Style-conditioned models prepend a \texttt{[style\_i]} prefix. Our importance-weighted objective uses the same training traces as the other methods; only the per-example loss weights differ. 

\paragraph{Baselines.}
Our primary comparison is \textsc{Vanilla SFT}, which uses the same teacher corpus without style prefixes or reweighting. We additionally compare against three controls.
\textsc{Style SFT (Empirical)} uses the discovered style prefixes while retaining the empirical training distribution.
\textsc{Random-Style SFT} replaces each discovered style prefix with one sampled uniformly at random from the six available styles, testing whether gains arise merely from introducing conditioning tokens.
\textsc{Teacher-Style SFT} uses teacher identity as the prefix, testing whether the discovered styles provide structure beyond the model that generated each trace. This model uses nine styles instead of six, one corresponding to each teacher.

\paragraph{Evaluation.}
We evaluate on MATH-500, AMC 12 (2025), OlympiadBench-EN, AIME 2024, AIME 2025, and HMMT February 2025. We sample with temperature $0.6$, top-$p=0.95$, and a maximum generation length of $4096$ tokens. For each style-conditioned model, we draw $256$ samples under each of the six style prefixes; teacher-conditioned models are evaluated analogously over nine teacher prefixes. Vanilla SFT and the base models produce $256$ unprefixed samples per problem. To compare methods at the same sampling budget, our reported mixed-style Pass@$k$ allocates the $k$ samples uniformly across the available style prefixes.
If $p_c$ is the empirical success probability under style $c$ and $k_c$ samples are allocated to that style, we compute
$
\mathrm{Pass@}k
=
1-
\prod_c
(1-p_c)^{k_c},
$
with the total budget satisfying
$
\sum_c k_c = k.
$
We report Pass@$k$ for $k$ up to 256.

\section{Results}
\subsection{Style-aware fine-tuning improves Pass@$k$}
\label{sec:passk-results}

Across benchmarks, \textsc{Style SFT (Importance-Weighted)} frequently improves substantially over both vanilla SFT and the base model, as shown in Figure~\ref{fig:06bpassk}. The same pattern holds for the 1.7B and 4B students. Full results for these models are reported in Appendix~\ref{app:fullresults}. In contrast, vanilla SFT often reduces Pass@$k$ relative to the base model. For the Qwen3 0.6B student, at $k=256$, vanilla SFT obtains $74.9\%$ on MATH-500, $46.6\%$ on AMC12-2025, and $44.2\%$ on OlympiadBench, while our method reaches $89.8\%$, $68.5\%$, and $60.1\%$, respectively. The difference is even more apparent on the harder competition benchmarks: Pass@$256$ increases from $9.7\%$ to $29.7\%$ on AIME 2024 and from $9.5\%$ to $30.1\%$ on AIME 2025. On HMMT, vanilla SFT reaches only $2.1\%$, compared with $11.4\%$ for our method.

The gains cannot be explained by the addition of style prefixes alone.
\textsc{Style SFT (Empirical)}, which uses the same discovered style labels while preserving their empirical frequencies, remains much closer to vanilla SFT. Likewise, replacing the discovered labels with random style prefixes does not reproduce the gains. These controls suggest that simply introducing discrete conditioning tokens is insufficient; the improvement emerges when style conditioning is combined with reweighting the contributions of the discovered styles during training.

\paragraph{Importance-weighted SFT improves style fidelity.}
The preceding results show that style prefixes affect performance, but this
alone does not establish that they induce the corresponding styles discovered
in the teacher traces. We measure whether generations conditioned
on \texttt{[style\_i]} fall into the same style mode as teacher traces assigned
to $s_i$. After generation, we assign each solution a realized style $\hat{s}$
using the nearest teacher-style centroid in our handcrafted style feature space (Table~\ref{tab:style-profile-full}) and compute
$P(\hat{s}=j\mid \texttt{[style\_i]})$.
For \textsc{Style SFT (Importance-Weighted)}, the fine-tuned models generally produce the style they were conditioned on: mean diagonal probability is
$30.8\%$, $30.8\%$, and $32.9\%$ for the 0.6B, 1.7B, and 4B students, respectively, compared with $16.7\%$ chance.
Style fidelity is concentrated in a subset of modes: $s_1$ and $s_3$ are recovered most reliably, while $s_2$, $s_4$, and $s_5$ are frequently realized as neighboring styles and $s_6$ shows moderate adherence.
In comparison, \textsc{Style SFT (Empirical)} remains close to chance
(Appendix~\ref{app:style-adherence}).
Thus, the learned prefixes provide control over
the discovered style structure, and the importance-weighted model exhibits
stronger style fidelity than conditioning on the empirical
distribution alone. Representative generations for each style prefix are shown in
Appendix~\ref{app:style-examples}.

\subsection{Styles induce problem-dependent solution distributions}
\label{sec:style-usefulness}

\begin{figure}[b]
    \centering
    \includegraphics[width=\linewidth]{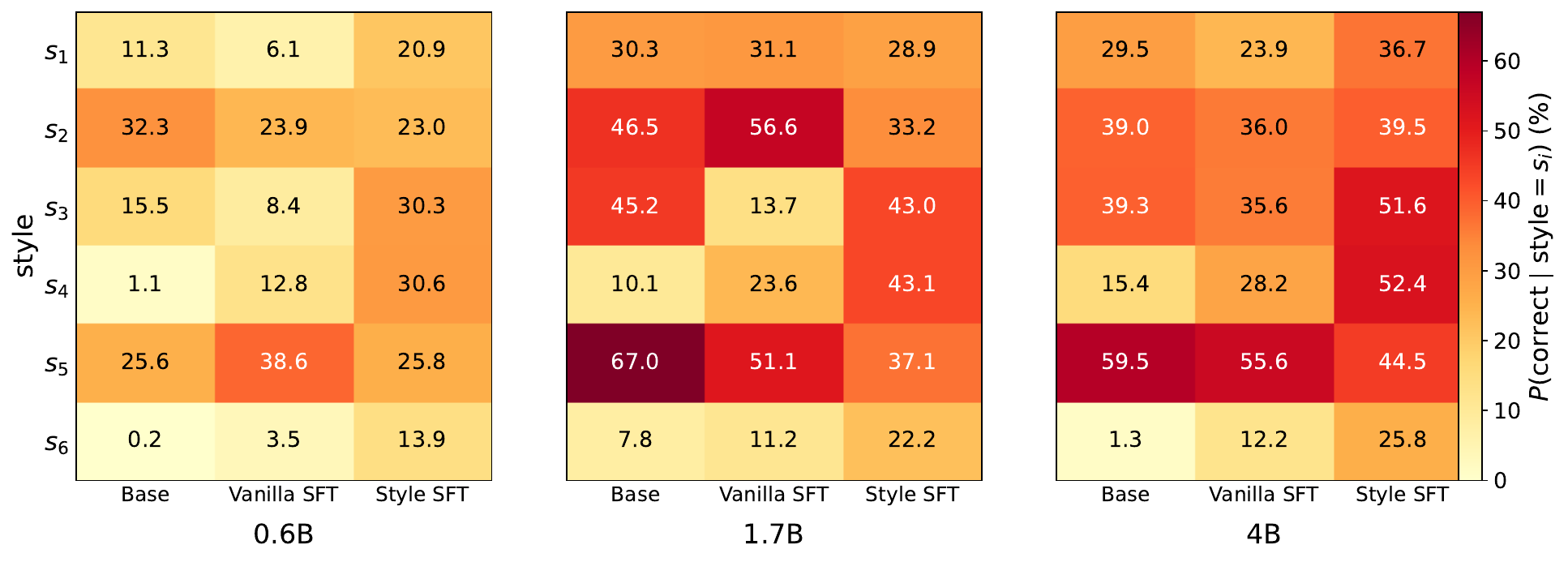}
    \caption{
    Correctness by style on MATH-500.
    Changing the style condition produces differences in accuracy at all three model scales.
    }
    \label{fig:style-conditioned-correctness}
\end{figure}

Having established that style-aware fine-tuning improves Pass@$k$, we next ask whether the discovered styles correspond to functionally different solution distributions. 
If style acted only as a surface-level control over
how an otherwise fixed solution is expressed, changing the style prefix should not substantially affect correctness. We find that this is not the case: style interacts with content, changing the correctness of the answers. We use MATH-500 for the following analyses.

\paragraph{Style conditioning changes solution accuracy.}
For the Style-SFT models trained with importance-weighted SFT, we generate
solutions under each of the six explicit style prefixes and measure
$P(\mathrm{correct}\mid s)$, so that style is fixed before generation.
As shown in Figure~\ref{fig:style-conditioned-correctness}, changing only the
style prefix produces differences in accuracy for the same trained
model. For comparison, the Base and Vanilla SFT models are decoded without
style prefixes, and their generations are assigned retrospectively to the
nearest teacher-style centroid using the handcrafted features from
Table~\ref{tab:style-profile-full}.

For the 0.6B Style SFT model, mean correctness ranges from $13.9\%$ under $s_6$ to
$30.6\%$ under $s_4$; for the 1.7B student, from $22.2\%$ to $43.1\%$; and
for the 4B student, from $25.8\%$ to $52.4\%$. The spread in correctness
across style conditions increases with model size: the difference between
the highest and lowest performing style is $16.7$ percentage points for
0.6B, $20.9$ points for 1.7B, and $26.6$ points for 4B. Thus, larger
students exhibit more differentiated style-conditioned solution distributions.

A second pattern is that several styles associated with relatively low
correctness under Base or Vanilla SFT become effective conditional modes under
Style SFT. For example, for the 0.6B model, generations retrospectively
assigned to $s_4$ have $12.8\%$ correctness under Vanilla SFT, compared with
$30.6\%$ when $s_4$ is explicitly requested from the Style-SFT model. At 4B,
the corresponding values are $28.2\%$ and $52.4\%$, with similar patterns for
$s_3$ and $s_6$. This analysis suggests that style-aware training makes several discovered styles available as useful controlled solution modes.

\paragraph{Style effects occur within individual problems.}
The aggregate differences above could still be explained by a global ordering in which some styles are simply more effective than others across all questions.
To test whether style instead interacts with problem content, we measure style sensitivity separately for each question.
Let
$
p_{x,s}
=
P(\mathrm{correct}\mid x,s),
$
and define
$
\Delta_x
=
\max_s (p_{x,s})
-
\frac{1}{6}\sum_{s=1}^{6}p_{x,s}.
$
Mean $\Delta_x$ is $0.104$, $0.125$, and $0.164$ for the 0.6B, 1.7B, and 4B students, respectively.
For the 0.6B student, $59\%$ of questions have a gap exceeding 5 percentage points and $45\%$ exceed 10 points; for 1.7B and 4B the 5 and 10 percentage point gaps rise to $74\%$ and $55\%$, and $84\%$ and $71\%$ respectively. 
Thus, even for a fixed problem, changing the style condition can change the probability of success.
This difference also appears in multi-sample performance: when all samples are drawn from a single style, the resulting Pass@$k$ curves differ across style conditions (Appendix~\ref{app:fixed-style-passk}). Thus, the choice of style also affects the coverage obtained from repeated sampling.

\paragraph{Different problems favor different styles.}
\begin{figure}[ht]
    \centering
    \includegraphics[width=\linewidth]{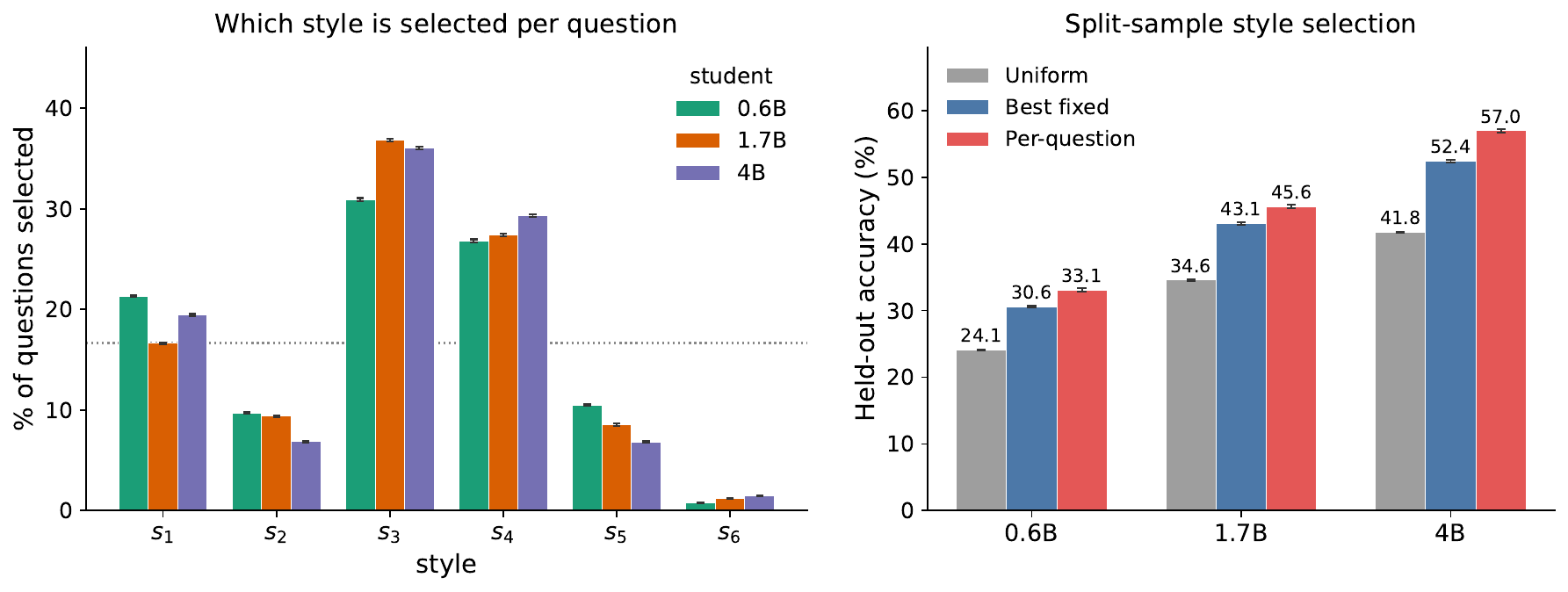}
    \caption{
    The most useful style depends on the problem.
    For each of 400 random splits, 128 of the 256 generations per question and style are used for style selection and the remaining 128 for held-out evaluation.
    Left: fraction of questions for which each style is selected as best; several different styles are preferred across problems.
    Right: held-out accuracy under a uniform mixture of styles, the globally best fixed style, and a style selected separately for each question.
    Question-specific selection consistently outperforms the globally best fixed style, showing that style usefulness cannot be explained by a single global ranking.
    }
    \label{fig:style-selection}
\end{figure}
Within-question variation alone does not rule out the possibility that one style is consistently best.
We test this with a split-sample style-selection experiment. For each question, we generate 256 answers in each of the six styles. For each style, we choose 128 answers to select the best style $s^\ast_x = \arg\max_s \hat{p}_{x,s}^{select}$ for that question. We then evaluate the remaining 128 generations conditioned on the selected style. We compare this to a globally best style, $
s_{\mathrm{global}}^*
=
\arg\max_s
\frac{1}{|\mathcal{X}|}
\sum_{x\in\mathcal{X}}
\hat{p}^{\mathrm{select}}_{x,s},$ determined by choosing the best style on average across questions. We repeat the procedure over 400 random splits of the 256 generated responses.

The selected styles vary substantially across questions.
Although $s_3$ and $s_4$ are chosen most often, styles $s_1$--$s_5$ are each preferred for a nontrivial subset of problems.
More importantly, this variation generalizes to held-out samples.
Per-question selection reaches $33.1\%$ accuracy for the 0.6B student compared with $30.6\%$ for the globally best fixed style; for 1.7B the corresponding values are $45.6\%$ and $43.1\%$, and for 4B they are $57.0\%$ and $52.4\%$.
The per-question advantage is positive at every model size, as shown in Figure~\ref{fig:style-selection}.

Taken together, these results show that the discovered styles correspond to functionally distinct conditional solution distributions. Changing the style condition alters the probability of success for the same problem, and this variation cannot be explained by a single global ranking of styles: different
problems favor different styles on held-out generations. This provides a possible explanation for why mixed-style sampling improves Pass@$k$: it gives the student access to multiple conditional solution behaviors whose usefulness varies across problems.

\section{Discussion}

\paragraph{Why does style reweighting help?}
Style prefixes alone provide relatively little improvement when the empirical training distribution is preserved, whereas importance-weighted style training produces substantially larger gains. One possible explanation is that rare styles receive too little supervision under the empirical distribution for the student to learn them as reliable conditional behaviors. Reweighting increases their contribution to the objective and may therefore make a broader set of style-conditioned solution distributions available at inference time. Importance-weighted training also produces substantially stronger correspondence between requested and realized styles than conditioning on the empirical distribution alone. The benefit of reweighting may come from preventing less common solution modes from being absorbed into the dominant behavior of the model.

More generally, these results suggest that diversity in a multi-teacher corpus is not necessarily preserved by standard training. A dataset may contain multiple useful ways of reasoning about the same problem, while their unequal frequency causes the student to preferentially learn only the dominant modes. Making this structure explicit and balancing its contribution to training provides one way to preserve more of the variation already present in the training data.

\paragraph{Style-conditioned behaviors become more differentiated with model size.}
The effect of style conditioning becomes stronger as student size increases. Both the gap between the highest and lowest performing style and the problem-level sensitivity to style increases with scale. 
On the other side, style fidelity itself does not increase nearly as sharply across model sizes: all three models exhibit comparable correspondence between requested and realized styles. 

One possibility is that larger models have sufficient capacity to represent several genuinely distinct conditional solution distributions, whereas smaller models partially collapse these conditions onto a more similar underlying behavior. Under this interpretation, the style prefix selects among increasingly differentiated reasoning behaviors as model capacity grows. Another possibility is that larger models can exploit the structural differences associated with each style more effectively: for example, patterns of verification, decomposition, or intermediate derivation may have little effect for a smaller student but become useful once the model is capable of carrying them out reliably. 

\paragraph{Why does correctness depend on style?}
Conditioning on different styles changes correctness for the same problem, and the most useful style varies across problems. Why this interaction arises remains an open question. One possibility is that a given model finds some forms of reasoning easier to generate or maintain than others. Styles that encourage particular forms of intermediate steps may make some solution trajectories easier to realize reliably. Another possibility is that in the benchmarks we used in evaluation, the problems themselves contain regularities that interact with these forms of reasoning, such that particular styles are better suited to particular classes of questions.

\section{Conclusion}
Current language models learn content and style jointly, making stylistic variation difficult to identify or control explicitly. In this work, we show that these sources of variation can be separated without style supervision, and that the resulting structure can be used to create more controllable, style-aware language models. We focus on mathematical reasoning and use an autoencoder to learn disentangled representations for the style and content of the problem solutions. We cluster the style representations and discover six styles in traces from nine models. Using these discovered styles, we fine-tune models that can be explicitly conditioned on style, by using an importance-weighted loss that encourages learning of all styles uniformly. Importance-weighted Style SFT improves Pass@$k$ over vanilla SFT across mathematical reasoning benchmarks and produces strong correspondence between requested and realized styles.

Our work has several limitations. The discovered styles depend on the representation-learning objective, the training corpus, and our discretization of the representation into clusters; other datasets or objectives may reveal different structure, and stylistic variation may ultimately be continuous or hierarchical rather than discrete. 
We also focus on mathematics because it allows both multiple solutions to a question and an objective way to evaluate correctness. We expect our approach for separating style and content to work in other domains that have these properties and leave it for future work to extend it to general text.
Moreover, we leave it for future work to study the cause of the interaction between style and correctness and figure out how to best use the learned styles at test time.

Overall, we identify style as a source of structure in data that standard next-token prediction does not represent explicitly. Instead, style is learned together with all other properties of the text, so standard model training simply inherits their distribution from the training traces and may underrepresent or collapse modes.
We show that we can discover these styles without labels, represent them explicitly, and control how they influence learning. Our work is a first step toward training style-aware language models, in which style is an explicit and controllable dimension of model behavior rather than an incidental byproduct of the data and training procedure.


\section*{Reproducibility statement}
The code is available at \url{https://github.com/ioanam25/reason-in-style}. The details for the autoencoder training are shown in Appendix~\ref{app:style-representation-details}. The fine-tuning implementation is described in Section~\ref{sec:style-aware-training}.

\section*{Acknowledgements}

We thank Sebastian Cygert for helpful discussions about this project.
This research was partly conducted as part of the Sovereign AI Foundation Model Project (GPU Track), organized by the Ministry of Science and ICT (MSIT) and supported by the National IT Industry Promotion Agency (NIPA), S.Korea. (PJT-26-010017), it was further supported by the MSIT (Ministry of Science, ICT), Korea, under the Top-Tier AI Global HRD invitation program (RS-2025-25461932) supervised by the IITP (Institute for Information \& Communications Technology Planning \& Evaluation) and the National Science Foundation (under NSF Award 1922658).

\bibliography{iclr2027_conference}
\bibliographystyle{iclr2027_conference}
\newpage
\appendix
\section{Style Representation Learning Details}
\label{app:style-representation-details}

This section gives the full objective and hyperparameters for the representation-learning method described in \cref{sec:method-ae}.

\subsection{Architecture}

Each minibatch contains $N$ questions, with two independently generated traces for each question, yielding $2N$ traces
\begin{equation}
\{s_{i,1},s_{i,2}\}_{i=1}^{N}.
\end{equation}

An encoder $E$ maps each trace $s$ to a representation $h(s)$. Two separate MLP projection heads produce a content representation $c(s)$ and a style representation $z(s)$:
$
c(s)\in\mathbb{R}^{d_c},
z(s)\in\mathbb{R}^{d_z},
$
where
$
d_c=d_z=256.
$

The two representations are concatenated and projected to a continuous prefix of
$
m=8
$
tokens. This prefix conditions an autoregressive decoder $D$, which reconstructs the original trace.

\subsection{Training objective}

The full objective is
\begin{equation}
\mathcal{L}
=
\lambda_c\mathcal{L}_c
+
\lambda_{\mathrm{orth}}\mathcal{L}_{\mathrm{orth}}
+
\lambda_{\mathrm{rec}}\mathcal{L}_{\mathrm{rec}}
+
\lambda_{\mathrm{var}}\mathcal{L}_{\mathrm{var}}
+
\lambda_{\mathrm{cov}}\mathcal{L}_{\mathrm{cov}}
+
\lambda_{\mathrm{div}}\mathcal{L}_{\mathrm{div}}.
\end{equation}

Unless otherwise specified, we use
\begin{equation}
\lambda_c
=
\lambda_{\mathrm{orth}}
=
\lambda_{\mathrm{rec}}
=
\lambda_{\mathrm{var}}
=
\lambda_{\mathrm{div}}
=
1,
\qquad
\lambda_{\mathrm{cov}}=0.15.
\end{equation}

The variance, covariance, and within-question diversity objectives operate only on $z$.

\paragraph{Content agreement.}

The two traces generated for the same question form a positive pair for the content representation. We normalize each content vector,
\begin{equation}
\bar{c}_i
=
\frac{c_i}{\|c_i\|},
\end{equation}
and define
\begin{equation}
\mathrm{sim}(i,j)
=
\frac{\bar{c}_i^\top \bar{c}_j}{\tau},
\qquad
\tau=0.07.
\end{equation}

For trace $i$, let $j^+(i)$ denote the other trace generated for the same question. All remaining traces in the minibatch serve as negatives. The content InfoNCE loss is
\begin{equation}
\mathcal{L}_c
=
-\frac{1}{2N}
\sum_{i=1}^{2N}
\log
\frac{
\exp\left(\mathrm{sim}(i,j^+(i))\right)
}{
\sum_{k\neq i}
\exp\left(\mathrm{sim}(i,k)\right)
}.
\end{equation}

This encourages $c$ to retain information that is stable across independently generated solutions to the same problem.

\paragraph{Content--style decorrelation.}

To discourage $c$ and $z$ from learning aligned representations of the same trace, we penalize their squared cosine similarity:
\begin{equation}
\mathcal{L}_{\mathrm{orth}}
=
\frac{1}{2N}
\sum_{r=1}^{2N}
\left(
\frac{
c(s_r)^\top z(s_r)
}{
\|c(s_r)\|\,\|z(s_r)\|+\epsilon
}
\right)^2,
\end{equation}
with
$
\epsilon=10^{-7}.
$

This encourages the two heads to occupy different directions in representation space, without requiring either representation to discard information needed for reconstruction.

\paragraph{Reconstruction.}

The reconstruction objective is the autoregressive cross-entropy of the decoder on the original trace:
\begin{equation}
\mathcal{L}_{\mathrm{rec}}
=
-\sum_t
\log
P_D
\left(
s_t
\mid
s_{<t},
c(s),
z(s)
\right).
\end{equation}

This ensures that the pair $[c(s);z(s)]$ remains jointly informative about the observed trace.

\paragraph{Variance regularization.}

To prevent collapse of the style representation, we use a VICReg-style variance constraint~\citep{bardes2022vicreg}. Let $\sigma_d(z)$ denote the standard deviation of the $d$-th dimension of $z$ across the $2N$ traces in a minibatch:
\begin{equation}
\sigma_d(z)
=
\sqrt{
\mathrm{Var}_{r=1,\ldots,2N}
\left[z(s_r)_d\right]
+
\varepsilon
},
\end{equation}
where $\varepsilon=10^{-4}$ is a small constant for numerical stability.

We impose a lower bound $\gamma=1$ on the standard deviation of each dimension:
\begin{equation}
\mathcal{L}_{\mathrm{var}}
=
\frac{1}{d_z}
\sum_{d=1}^{d_z}
\max\left(
0,\,
\gamma-\sigma_d(z)
\right).
\end{equation}

This penalizes dimensions of $z$ whose variation across the minibatch falls below $\gamma$, preventing the style representation from collapsing to a constant or near-constant representation.

\paragraph{Covariance regularization.}

Variance regularization prevents individual dimensions from collapsing but does not prevent them from becoming redundant. We therefore penalize the off-diagonal entries of the centered covariance matrix of $z$:
\begin{equation}
\mathcal{L}_{\mathrm{cov}}
=
\frac{1}{d_z}
\sum_{i\neq j}
\mathrm{Cov}(z)_{ij}^{2}.
\end{equation}

This follows the redundancy-reduction principle used by VICReg and related methods~\citep{bardes2022vicreg,zbontar2021barlow}, encouraging variation in $z$ to be distributed across multiple dimensions.

\paragraph{Within-question style diversity.}

Finally, we prevent $z$ from inheriting the same same-question invariance imposed on $c$. For question $i$, let
\begin{equation}
\rho_i
=
\frac{
z(s_{i,1})^\top z(s_{i,2})
}{
\|z(s_{i,1})\|\,\|z(s_{i,2})\|
}.
\end{equation}

We penalize positive similarity between the two style representations:
\begin{equation}
\mathcal{L}_{\mathrm{div}}
=
\frac{1}{N}
\sum_{i=1}^{N}
\mathrm{ReLU}(\rho_i).
\end{equation}

Thus, the content objective encourages same-question solutions to agree in $c$, while the diversity objective prevents the corresponding $z$ representations from collapsing together. Combined with reconstruction, this encourages $z$ to retain trace-specific variation that is complementary to the information represented by $c$.

\subsection{Implementation Details}
\label{app:style-implementation}

\paragraph{Encoder and decoder.}
We use \texttt{Alibaba-NLP/gte-Qwen2-1.5B-instruct} as the encoder.
The pretrained encoder parameters are frozen and adapted with LoRA of rank $8$,
scaling parameter $\alpha=16$, and dropout $0.05$ on the query, key, value,
and output projections.
We obtain the trace representation using last-token pooling.
The encoder uses scaled dot-product attention and gradient checkpointing.

For reconstruction, we use the post-trained \texttt{Qwen/Qwen3-4B} model as
the decoder unless noted otherwise.
The decoder is adapted with LoRA of rank $16$, $\alpha=32$, and dropout $0.05$
on the query, key, value, and output projections.
The decoder's base parameters remain frozen, while the LoRA parameters and
the continuous-prefix projection are trained.

\paragraph{Projection heads.}
The content and style heads have identical architectures.
Starting from the $1536$-dimensional pooled encoder representation, each head
is a four-layer MLP:
\begin{equation}
\begin{aligned}
\mathrm{LayerNorm}(1536)
&\rightarrow \mathrm{Linear}(1536,256)
\rightarrow \mathrm{GELU}
\rightarrow \mathrm{Linear}(256,256)
\rightarrow \mathrm{GELU}
\\
&\rightarrow \mathrm{Linear}(256,256)
\rightarrow \mathrm{GELU}
\rightarrow \mathrm{Linear}(256,256).
\end{aligned}
\end{equation}
There is no final activation or normalization in either head.
For the InfoNCE objective, $c$ is normalized explicitly when computing cosine
similarity.

\paragraph{Continuous reconstruction prefix.}
The concatenated representation
$
[c(s);z(s)]\in\mathbb{R}^{512}
$
is mapped by a single learned linear projection to
$
8\times 2560
$
values and reshaped into $m=8$ continuous embeddings in the decoder's input
space.
These embeddings are prepended to the decoder input through
\texttt{inputs\_embeds}.
The encoder and decoder use their respective tokenizers.
When required by the decoder tokenizer, a beginning-of-sequence token is
inserted after the continuous prefix and before the trace tokens.
The reconstruction cross-entropy is computed only over the original trace
tokens, excluding the continuous prefix.

\paragraph{Optimization.}
We train for four epochs using 8 GPUs with distributed data parallelism and
\texttt{bf16} mixed precision.
The maximum sequence length is $4096$ tokens.
Each GPU processes one pair of traces from the same question per forward
pass, giving 8 question pairs, or $2N=16$ traces, across devices.
Representations are gathered across devices before computing the
batch-dependent objectives.
We use gradient accumulation over two forward passes, corresponding to an
effective optimizer batch of 16 question pairs.

We optimize with AdamW using
\begin{equation}
\beta_1=0.9,
\qquad
\beta_2=0.999,
\qquad
\mathrm{weight\ decay}=0.01.
\end{equation}
The content and style heads, decoder LoRA parameters, and prefix projection
use a learning rate of $10^{-4}$, while the encoder LoRA parameters use a
learning rate of $10^{-6}$.
We use a constant learning-rate schedule with no warmup and clip the gradient
norm at $10$.


\section{Controlled Style Recovery}
\label{app:known-styles}

This appendix provides the full setup and ablations for the controlled style-recovery experiment summarized in \cref{sec:known-styles}. The goal is to verify that the proposed representation can recover a known stylistic factor without using style labels during training, and that stylistic information is isolated in $z$ rather than duplicated across both representations.

\subsection{Dataset and prompted styles}

We prompt Gemini 3.1 Pro to solve 300 mathematical problems from the DeepScaleR dataset in four explicitly specified writing styles, yielding 1{,}200 traces in total. Each problem is generated in all four styles, ensuring that style is not confounded with question identity. We use a question-level 1{,}080/120 train/test split, so no question appears in both sets. Style labels are used only for evaluation.

The four prompted styles are shown in \cref{tab:gemini-styles}. They are deliberately chosen to differ strongly in observable surface form, providing a controlled positive test of whether the representation can separate question-specific content from stylistic variation.

\begin{table}[ht]
\centering
\caption{Four prompted Gemini writing styles used in the controlled experiment.}
\label{tab:gemini-styles}
\scriptsize
\setlength{\tabcolsep}{4pt}
\begin{tabular}{clp{7.0cm}}
\toprule
ID & Name & Prompted surface form \\
\midrule
0 & Academic
& Formal paragraphs; passive voice; frequent use of connective phrases such as ``therefore'' and ``it follows that''. \\
1 & ELI5 / tutor
& Short sentences; explanatory analogies; encouraging, tutorial-like tone. \\
2 & Algorithmic list
& Step-by-step organization, with each sentence explicitly formatted as a numbered or ordered step. \\
3 & Pure equation
& Minimal natural language; equation-dominated derivations using symbols such as $\Rightarrow$ and $\therefore$. \\
\bottomrule
\end{tabular}
\end{table}

\subsection{Model and training details}

The encoder is GTE--Qwen2-1.5B-Instruct with LoRA rank $r{=}8$. The decoder is SmolLM2-135M~\citep{allal2025smollm2smolgoesbig}, frozen except for LoRA adapters with rank $r{=}16$. 


We compare three representations or training objectives:

\begin{itemize}
    \item \textbf{Frozen encoder $h$:} the off-the-shelf GTE representation, without LoRA adaptation or a $c/z$ split.
    \item \textbf{Reconstruction-only:} the same autoencoder architecture trained with $\lambda_{\mathrm{rec}}=1$ and all auxiliary loss weights set to zero.
    \item \textbf{Full objective:} the complete content--style disentangling objective described in \cref{sec:method-ae}.
\end{itemize}

\subsection{Evaluation}

We evaluate whether the learned representations organize according to question identity and prompted style.

\paragraph{Cosine-similarity gap.}
For a given grouping variable, we compute the mean cosine similarity between pairs sharing the same label and subtract the mean similarity between pairs with different labels. Higher values indicate stronger clustering by the corresponding attribute.

\paragraph{Unsupervised clustering.}
We apply $k$-means with $k{=}4$ and compare the resulting partition against the ground-truth style labels using adjusted Rand index (ARI) and normalized mutual information (NMI).

\paragraph{Linear probing.}
We train a 4-way linear classifier to predict the ground-truth style ID. Chance accuracy is $25\%$. We additionally evaluate a question-held-out probe in which the classifier is trained and evaluated on disjoint sets of questions.

\subsection{Full disentanglement results}

\cref{tab:gemini-disentangle} reports the complete results. Under the full objective, the content representation $c$ strongly organizes traces by question, with a question cosine gap of $0.93$, while showing essentially no clustering by style. In contrast, the style representation $z$ has a style gap of $1.18$ and a negative question gap of $-0.30$.

Style can also be recovered directly from $z$. A linear probe predicts the prompted style with $96.7\%$ accuracy, while the same probe on $c$ achieves only $23.3\%$, close to the $25\%$ chance level. Similarly, $k$-means on $z$ recovers the four prompted styles with ARI $0.77$ and NMI $0.78$, whereas clustering $c$ gives essentially no agreement with style.

\begin{table}[t]
\centering
\caption{
Full disentanglement results on the controlled Gemini setting.
Gap is the mean within-group cosine similarity minus the mean between-group similarity.
ARI/NMI are from $k$-means with $k{=}4$ against the ground-truth \texttt{style\_id}.
The style probe is a 4-way linear classifier with chance accuracy $25\%$.
}
\label{tab:gemini-disentangle}
\scriptsize
\setlength{\tabcolsep}{4pt}
\resizebox{0.6\linewidth}{!}{%
\begin{tabular}{lccc}
\toprule
Metric & Frozen GTE $h$ & Recon-only & Full objective \\
\midrule
$c$ question gap & --- & 0.36 & \textbf{0.93} \\
$c$ style gap & --- & 0.09 & $\mathbf{\approx 0}$ \\
$z$ style gap & --- & 0.10 & \textbf{1.18} \\
$z$ question gap & --- & 0.38 & $\mathbf{-0.30}$ \\
$z$ style ARI / NMI & 0.29 / 0.36 & 0.85 / 0.82 & 0.77 / 0.78 \\
$c$ style ARI / NMI & --- & 0.91 / 0.87 & $\mathbf{\approx 0}$ / $0.00$ \\
$z$ style probe & 96.7\% & 96.3\% & 96.7\% \\
$c$ style probe & --- & 97.1\% & \textbf{23.3\%} \\
\bottomrule
\end{tabular}%
}
\end{table}

\paragraph{Why the auxiliary objectives matter}

Reconstruction alone is sufficient to preserve substantial stylistic information, but it does not separate that information from content. Under the reconstruction-only objective, both representations strongly encode style: $z$ achieves style ARI $0.85$ and probe accuracy $96.3\%$, while $c$ achieves style ARI $0.91$ and probe accuracy $97.1\%$.

Thus, successful reconstruction does not by itself induce a content--style factorization. Without the auxiliary objectives, the same stylistic information is duplicated across both heads. In contrast, the full objective retains style information in $z$ while removing it from $c$, reducing the style ARI of $c$ to approximately zero and its probe accuracy to chance.

This distinction is important for the downstream use of the representation: our goal is not merely to encode style somewhere in the latent space, but to obtain a representation in which recurring stylistic variation can be isolated and clustered independently of question-specific content.

\paragraph{Comparison with the pretrained encoder}

The off-the-shelf encoder already contains substantial information about prompted style. A supervised linear probe predicts the style label from the frozen representation $h$ with $96.7\%$ accuracy. However, this information is not naturally organized into the four style modes: unsupervised $k$-means on $h$ obtains only ARI $0.29$ and NMI $0.36$.

This comparison separates \emph{encoding} style information from \emph{organizing} the representation around style. The pretrained encoder makes style linearly accessible, but style labels are required to read it out. The learned $z$ representation instead makes the same structure recoverable without supervision.

\paragraph{Generalization across questions}

We additionally test whether the learned style structure generalizes to unseen problem instances. A linear style classifier trained on one set of questions predicts style on held-out questions with $95.8\%$ accuracy. This indicates that the information captured in $z$ is not tied to particular mathematical problems, but instead reflects recurring variation that transfers across questions.

Overall, the controlled experiment establishes three properties needed for the main analysis: the proposed representation recovers a known stylistic factor without using its labels, style is isolated in $z$ rather than leaking into $c$, and the resulting structure generalizes across questions. These results motivate using $z$ to discover recurring styles in the unlabeled multi-teacher reasoning corpus.

\section{Teacher Styles}
\label{app:teacher-styles}

This appendix provides additional details on the six styles discovered in the multi-teacher SCAS corpus, including their relationship to teacher identity, question identity, and handcrafted surface features.

\subsection{Style discovery setup}
\label{app:style-discovery-setup}

We use the subset of the SCAS verified teacher pool containing 12{,}426 questions with one solution from each of nine teacher models, yielding 111{,}834 traces in total with maximum length 4096 tokens. The nine teachers are
\texttt{gemma-4-31b-it}~\citep{gemma4techreport},
\texttt{gpt-5-chat}~\citep{openai2025gpt5},
\texttt{gpt-oss-120b}~\citep{openai2025gptoss},
\texttt{llama-3.3-70b-instruct}~\citep{dubey2024llama3},
\texttt{olmo-3.1-32b-instruct}~\citep{teamolmo2025olmo3},
\texttt{phi-4-reasoning-plus}~\citep{abdin2025phi4reasoning},
\texttt{qwen2.5-72b-instruct}~\citep{yang2024qwen25},
\texttt{qwen3-32b}~\citep{qwenteam2025qwen3}, and
\texttt{qwen3.5-27b}~\citep{qwenteam2026qwen35}. We split this dataset into 90\% train and 10\% validation.

The questions are drawn from MATH~\citep{hendrycks2021math}, GSM8K~\citep{cobbe2021gsm8k}, DeepScaleR~\citep{deepscaler2025}, and OpenR1-Math~\citep{openr12025}. The dataset is available at \url{https://huggingface.co/datasets/Student-Centric-Answer-Sampling/scas_verified_teacher_pool}.

After training the style autoencoder, we encode each trace and fit a full-covariance Gaussian mixture model to $z(s)$. We use $K{=}6$ for the experiments, chosen by best Pass@$k$ on the validation set. The clusters are ordered by mean trace length and denoted $s_1,\ldots,s_6$, from shortest to longest.


\subsection{Analysis of Discovered Teacher Styles}
\paragraph{Teacher preferences over styles.}
\label{app:teacher-style-composition}

Teachers differ substantially in how frequently they generate the six styles. \cref{tab:teacher-clusters-full} reports $P(s\mid\text{teacher})$. Llama, gemma, and qwen2.5 place substantial mass on the short textbook style $s_1$, whereas phi-4, qwen3, and qwen3.5 generate many more long think-block traces in $s_6$. The dominant $s_5$ cluster is comparatively common across all teachers. Despite these preferences, style is not equivalent to teacher identity. The mutual information between GMM style and teacher ID is $0.33$ nats, corresponding to normalized mutual information $0.17$.

\begin{table}[ht]
\centering
\caption{
Share of each teacher's traces assigned to each $K{=}6$ style (\%; rows sum to 100).
The final row gives the overall style distribution across all 111{,}834 traces.
}
\label{tab:teacher-clusters-full}
\scriptsize
\setlength{\tabcolsep}{2.5pt}
\resizebox{0.5\linewidth}{!}{%
\begin{tabular}{lcccccc}
\toprule
Teacher & $s_1$ & $s_2$ & $s_3$ & $s_4$ & $s_5$ & $s_6$ \\
\midrule
gemma-4-31b-it          & 41.6 & 16.1 & 0.7  & 2.3  & 37.0 & 2.3 \\
gpt-5-chat              & 0.4  & 10.6 & 23.9 & 22.1 & 37.1 & 5.9 \\
gpt-oss-120b            & 0.1  & 5.4  & 28.1 & 21.4 & 44.3 & 0.8 \\
llama-3.3-70b-instruct  & 47.6 & 15.7 & 0.0  & 0.1  & 32.4 & 4.2 \\
olmo-3.1-32b-instruct   & 5.6  & 12.1 & 15.2 & 17.7 & 44.7 & 4.7 \\
phi-4-reasoning-plus    & 0.0  & 8.0  & 7.5  & 17.2 & 29.4 & 37.8 \\
qwen2.5-72b-instruct    & 39.5 & 13.0 & 2.1  & 4.8  & 40.5 & 0.1 \\
qwen3-32b               & 0.6  & 6.4  & 0.3  & 2.4  & 61.6 & 28.8 \\
qwen3.5-27b             & 2.6  & 10.8 & 3.2  & 4.1  & 45.0 & 34.3 \\
\midrule
All traces               & 15.3 & 10.9 & 9.0  & 10.2 & 41.3 & 13.2 \\
\bottomrule
\end{tabular}%
}
\end{table}

\paragraph{Styles differ from question identity.}
\label{app:style-vs-question}

A potential concern is that the clusters reflect properties of the mathematical problem rather than properties of the generated solution. For example, difficult questions might systematically elicit longer traces, causing all teachers to be assigned to the same cluster. The structure of the dataset allows us to test this directly. For every question, we observe nine independent teacher solutions. Only $0.85\%$ of questions have all nine traces assigned to the same style. The average question contains $3.38$ distinct styles, and $83\%$ of questions contain at least three distinct styles.

\begin{table}[t]
\centering
\caption{
Number of distinct styles among the nine teacher solutions to the same question.
There are 12{,}426 aligned questions with nine traces each.
}
\label{tab:style-vs-question-full}
\scriptsize
\setlength{\tabcolsep}{5pt}
\begin{tabular}{lcccccc}
\toprule
\# distinct styles & 1 & 2 & 3 & 4 & 5 & 6 \\
\midrule
\% of questions
& 0.8 & 15.8 & 36.6 & 38.3 & 8.2 & 0.3 \\
\# questions
& 105 & 1{,}963 & 4{,}545 & 4{,}754 & 1{,}021 & 38 \\
\bottomrule
\end{tabular}
\end{table}



\paragraph{Observable properties of the discovered styles.}
\label{app:style-surface-features}

To characterize the discovered clusters without using teacher identity, we compute a small set of handcrafted trace-level features inspired by \citet{lippmann2025stylesubstance}. These include trace length, formatting statistics, and discourse markers:

\begin{itemize}
    \item number of words;
    \item number of lines per 100 words;
    \item equals-sign density;
    \item density of backtracking or ``wait'' language; and
    \item density of verification or checking language.
\end{itemize}

Except for total word count, density features are measured per 100 words. For each feature, we report the mean within each style and
\begin{equation}
\eta^2
=
\frac{\mathrm{SS}_{\mathrm{between\ styles}}}
{\mathrm{SS}_{\mathrm{total}}},
\end{equation}
the fraction of feature variance explained by style membership.

\begin{table}[ht]
\centering
\caption{
Handcrafted properties of the six discovered styles.
$\eta^2$ gives the fraction of variance in each feature explained by style membership.
}
\label{tab:style-profile-full}
\scriptsize
\setlength{\tabcolsep}{3pt}
\begin{tabular}{lccccccc}
\toprule
Feature & $s_1$ & $s_2$ & $s_3$ & $s_4$ & $s_5$ & $s_6$ & $\eta^2$ \\
\midrule
$n$ words           & 311  & 330  & 417  & 483  & 486  & 723  & 0.09 \\
dens.\ backtrack    & 0.01 & 0.10 & 0.12 & 0.20 & 0.17 & 0.34 & 0.07 \\
dens.\ verification & 0.02 & 0.08 & 0.04 & 0.07 & 0.11 & 0.20 & 0.09 \\
dens.\ equals       & 6.2  & 4.7  & 8.4  & 9.3  & 7.0  & 5.9  & 0.06 \\
lines / 100w        & 13.0 & 17.5 & 27.9 & 26.5 & 19.9 & 13.8 & 0.12 \\
\bottomrule
\end{tabular}
\end{table}

The styles are ordered by length, but length alone does not characterize the partition. Formatting is the strongest of the measured separators, with $\eta^2=0.12$ for lines per 100 words, while backtracking and verification also differ systematically. At the same time, all effect sizes remain moderate ($\eta^2\approx0.06$--$0.12$), indicating substantial variation within each cluster.


\paragraph{Summary.}

The discovered styles exhibit three properties relevant to our downstream use of them. First, they correspond to systematic differences in observable reasoning-trace form. Second, they recur across teacher models and across solutions to the same mathematical problem, rather than simply encoding teacher or question identity. Third, their empirical frequencies are highly imbalanced, with the dominant style accounting for more than $40\%$ of the training traces. These properties motivate using the discovered styles as a basis for controlling the composition of the distillation data.

\subsection{Examples of Discovered Styles in Teacher Traces}
\label{app:teacher-style-examples}

To illustrate the styles discovered in the original teacher corpus, we show
solutions to the same mathematical problem assigned to each of the six GMM
clusters. Holding the question fixed makes differences in organization,
verbosity, formatting, and reasoning presentation easier to compare.

The examples are drawn from the SCAS training corpus and labeled using the
style representation and $K=6$ GMM described in
Section~\ref{sec:scas-styles}. The style names are descriptive glosses assigned
after clustering and are not provided to the teacher models.

\paragraph{Question.}
For how many integer values of $n$ between 1 and 1000 inclusive does the decimal representation of $\frac{n}{1400}$ terminate?

Gold answer: $142$. Style counts among the nine teachers: style\_1: 2, style\_2: 2, style\_3: 1, style\_4: 1, style\_5: 2, style\_6: 1.

\paragraph{Style 1 (short textbook).}
2 teacher traces were assigned to this cluster; we show all of them.

\smallskip\noindent\textit{Teacher:} \texttt{qwen2.5-72b-instruct}

\begin{quote}
\small
To determine for how many integer values of \( n \) between 1 and 1000 inclusive the decimal representation of \(\frac{n}{1400}\) terminates, we need to understand when a fraction has a terminating decimal. A fraction in its simplest form has a terminating decimal if and only if the denominator (in its simplest form) has no prime factors other than 2 and 5.

First, we factorize the denominator 1400:
\[
1400 = 14 \times 100 = (2 \times 7) \times (2^2 \times 5^2) = 2^3 \times 5^2 \times 7
\]
The prime factorization of 1400 includes the prime number 7, which means that \(\frac{n}{1400}\) will not have a terminating decimal unless the 7 in the denominator is canceled out by a factor of 7 in the numerator \( n \).

Thus, \( n \) must be a multiple of 7. We need to find how many multiples of 7 there are between 1 and 1000 inclusive. This is equivalent to finding the largest integer \( k \) such that \( 7k \leq 1000 \):
\[
k = \left\lfloor \frac{1000}{7} \right\rfloor = \left\lfloor 142.857 \right\rfloor = 142
\]
So, there are 142 multiples of 7 between 1 and 1000. Therefore, there are 142 integer values of \( n \) for which the decimal representation of \(\frac{n}{1400}\) terminates.

The final answer is:
\[
\boxed{142}
\]
\end{quote}

\smallskip\noindent\textit{Teacher:} \texttt{olmo-3.1-32b-instruct}

\begin{quote}
\small
To determine for how many integer values of \( n \) between 1 and 1000 inclusive the decimal representation of \( \frac{n}{1400} \) terminates, we need to recall the condition for a fraction to have a terminating decimal expansion.

A rational number \( \frac{a}{b} \) in lowest terms has a terminating decimal if and only if the denominator (after simplifying the fraction) has no prime factors other than 2 or 5. In other words, the denominator's prime factorization contains only the primes 2 and/or 5.

Let's factorize 1400:

\[
1400 = 14 \times 100 = (2 \times 7) \times (2^2 \times 5^2) = 2^3 \times 5^2 \times 7
\]

So the prime factorization is \( 2^3 \times 5^2 \times 7 \).

For \( \frac{n}{1400} \) to have a terminating decimal, after simplifying the fraction, the denominator should have only 2s and 5s as prime factors. That means the factor of 7 in the denominator must be canceled out by a factor of 7 in the numerator \( n \).

Let's formalize this:

Let \( d = \gcd(n, 1400) \). Then

\[
\frac{n}{1400} = \frac{n/d}{1400/d}
\]

For the denominator \( 1400/d \) to have no prime factors other than 2 and 5, all factors of 7 in 1400 must be canceled by \( d \). That is, \( d \) must be divisible by 7.

But more precisely, since 1400 has one factor of 7, in order for \( 1400/d \) to not have a factor of 7, \( d \) must include that 7, i.e., 7 divides \( d \).

But actually, let's think step by step:

Let's write \( n = 7k \cdot m \), where \( k \) is an integer and \( m \) is not divisible by 7 (i.e., \( k \) is the highest power of 7 dividing n). But perhaps a better way is:

Let's let \( d = \gcd(n, 1400) \). Then the reduced denominator is \( \frac{1400}{d} \). We want \( \frac{1400}{d} \) to have no factor of 7, i.e., all the 7s in 1400 must be canceled by d. Since 1400 has exactly one factor of 7, d must be divisible by 7.

That is, 7 divides d, i.e., 7 divides \( \gcd(n, 1400) \), i.e., 7 divides n.

Thus, n must be divisible by 7.

But wait: Let's check this more carefully.

Let's suppose n is divisible by 7, say n = 7m.

Then:

\[
\frac{n}{1400} = \frac{7m}{1400} = \frac{m}{200}
\]

Now, 200 factors as \( 2^3 \times 5^2 \), which contains only 2s and 5s. So any fraction with denominator 200 (in lowest terms or not) will have a terminating decimal.

But wait, what if m and 200 have common factors? That's fine, because after simplifying, the denominator will still have only 2s and 5s as factors.

So if n is divisible by 7, then the reduced denominator is a divisor of 200, which is only 2s and 5s.

Now, what if n is NOT divisible by 7?

Then, in the reduced form, the denominator will still have a factor of 7 (since the numerator can't cancel it), so the decimal will not terminate.

Therefore, the decimal expansion of \( \frac{n}{1400} \) terminates if and only if n is divisible by 7.

Now we just need to count the number of integers n between 1 and 1000 inclusive that are divisible by 7.

Let's compute:

The smallest such n is 7, the largest is the greatest multiple of 7 less than or equal to 1000.

Compute \( \left\lfloor \frac{1000}{7} \right\rfloor = 142 \) since \( 7 \times 142 = 994 \), and \( 7 \times 143 = 1001 > 1000 \).

Thus, the multiples of 7 between 1 and 1000 are:

7 \(\times\) 1, 7 \(\times\) 2, ..., 7 \(\times\) 142

So there are 142 such integers.

Let's check with a quick example:

For n = 7: \( \frac{7}{1400} = \frac{1}{200} = 0.005 \) (terminates) For n = 14: \( \frac{14}{1400} = \frac{1}{100} = 0.01 \) (terminates) For n = 1: \( \frac{1}{1400} \) --- denominator has 7, so decimal does not terminate.

Thus, the answer is 142.

But wait, let's consider if there are any other restrictions. For example, what if n is divisible by higher powers of 7? But as shown above, even if n is divisible by 49 or 343, after dividing by 7, the denominator becomes 200, and any further common factors only reduce the denominator further, but never introduce new primes.

So the answer is correct.

\textbf{Final Answer:} \boxed{142}
\end{quote}

\paragraph{Style 2 (short unmarked).}
2 teacher traces were assigned to this cluster; we show all of them.

\smallskip\noindent\textit{Teacher:} \texttt{llama-3.3-70b-instruct}

\begin{quote}
\small
\textbf{Step 1: To determine if the decimal representation of }$\frac{n}{1400}$ terminates, we need to analyze the prime factors of the denominator. The prime factorization of 1400 is $2^3 \cdot 5^2 \cdot 7$. For the decimal representation to terminate, the denominator must only have prime factors of 2 and 5 after simplification with the numerator.

\textbf{Step 2: We need to find the values of }$n$ that will simplify the factor of 7 in the denominator. This means $n$ must be a multiple of 7 for the fraction $\frac{n}{1400}$ to simplify to a form where the denominator only contains factors of 2 and 5.

\textbf{Step 3: Calculate the number of multiples of 7 between 1 and 1000 inclusive.} To do this, we divide 1000 by 7 and take the floor of the result, which gives us the number of multiples of 7 in the range.

\textbf{Step 4: Perform the calculation.} $\left\lfloor\frac{1000}{7}\right\rfloor = \left\lfloor 142.857 \right\rfloor = 142$.

The final answer is: $\boxed{142}$
\end{quote}

\smallskip\noindent\textit{Teacher:} \texttt{gemma-4-31b-it}

\begin{quote}
\small
To determine for how many integer values of $n$ between 1 and 1000 inclusive the decimal representation of $\frac{n}{1400}$ terminates, we follow these steps:

1.  \textbf{Condition for a terminating decimal:}     A simplified fraction $\frac{a}{b}$ (where $\gcd(a, b) = 1$) has a terminating decimal representation if and only if the prime factorization of the denominator $b$ contains no prime factors other than 2 and 5.

2.  \textbf{Analyze the denominator:}     The denominator is $1400$. We find its prime factorization:     $1400 = 14 \times 100 = (2 \times 7) \times (10^2) = 2 \times 7 \times (2 \times 5)^2 = 2 \times 7 \times 2^2 \times 5^2 = 2^3 \times 5^2 \times 7$.

3.  **Determine the condition for $n$:**     The fraction is $\frac{n}{1400} = \frac{n}{2^3 \times 5^2 \times 7}$.     For the decimal to terminate, the prime factor $7$ in the denominator must be canceled out by a factor of $7$ in the numerator $n$.      Therefore, $n$ must be a multiple of 7.

4.  \textbf{Count the multiples of 7 between 1 and 1000:}     We need to find the number of integers $n$ such that $1 \le n \le 1000$ and $n$ is a multiple of 7.     This is equivalent to finding the largest integer $k$ such that $7k \le 1000$.     Dividing 1000 by 7:     $1000 \div 7 = 142.857...$

The largest integer $k$ is $\lfloor \frac{1000}{7} \rfloor = 142$.

The multiples of 7 are $7 \times 1, 7 \times 2, \dots, 7 \times 142$. There are 142 such values.
\end{quote}

\paragraph{Style 3 (contest markdown).}
One teacher trace was assigned to this cluster.

\smallskip\noindent\textit{Teacher:} \texttt{gpt-5-chat\_2025-10-03}

\begin{quote}
\small
Let's restate the problem carefully.

We want the number of integer values \( n \) between 1 and 1000 (inclusive) for which the decimal representation of

\[
\frac{n}{1400}
\]

\textbf{terminates}.

---

\textbf{Step 1. Recall the condition for a terminating decimal.}

A fraction \( \frac{a}{b} \) in lowest terms has a terminating decimal if and only if the denominator \( b \) has no prime factors other than 2 or 5.

That is, after simplifying \( \frac{a}{b} \), the denominator must be of the form \( 2^x 5^y \).

---

\textbf{Step 2. Factorize the denominator }\( 1400 \).

\[
1400 = 14 \times 100 = (2 \times 7) \times (2^2 \times 5^2) = 2^3 \times 5^2 \times 7
\]

So \( 1400 = 2^3 \cdot 5^2 \cdot 7 \).

---

\textbf{Step 3. Simplify the fraction }\( \frac{n}{1400} \).

When we reduce \( \frac{n}{1400} \) to lowest terms, we can cancel out common factors of 2, 5, or 7 that occur in \( n \).

Let's write

\[
n = 2^a \cdot 5^b \cdot 7^c \cdot k
\]

where \( k \) is not divisible by 2, 5, or 7.

Then

\[
\frac{n}{1400} = \frac{2^a \cdot 5^b \cdot 7^c \cdot k}{2^3 \cdot 5^2 \cdot 7}
= \frac{k}{2^{3-a} \cdot 5^{2-b} \cdot 7^{1-c}}
\]
after canceling common factors (some exponents may become 0 or negative, meaning those primes disappear from the denominator).

---

\textbf{Step 4. Condition for termination.}

For the decimal to terminate, the denominator after simplification must have *no factor of 7* (and only 2's and 5's allowed).

That means the factor of 7 in the denominator must cancel completely.

So we need \( c \ge 1 \), i.e. \( n \) must be \textbf{divisible by 7}.

---

\textbf{Step 5. That's the only requirement.}

If \( n \) is divisible by 7, then the denominator after cancellation will be \( 2^x 5^y \), which gives a terminating decimal.

If \( n \) is \textbf{not} divisible by 7, then a factor of 7 remains in the denominator, making the decimal \textbf{non-terminating}.

---

\textbf{Step 6. Count how many }\( n \) between 1 and 1000 are multiples of 7.

We need integers \( n \) such that \( n = 7k \) and \( 1 \le n \le 1000 \).

The largest \( k \) satisfying \( 7k \le 1000 \) is \( k = \lfloor 1000 / 7 \rfloor = 142 \).

So the positive multiples of 7 up to 1000 are \( 7, 14, \dots, 994 \). There are \textbf{142} of them.

---

\textbf{Final Answer:}

\[
\boxed{142}
\]
\end{quote}

\paragraph{Style 4 (longer markdown).}
One teacher trace was assigned to this cluster.

\smallskip\noindent\textit{Teacher:} \texttt{gpt-oss-120b}

\begin{quote}
\small
The fraction \(\dfrac{n}{1400}\) will have a terminating decimal exactly when, after reducing the fraction to lowest terms, the denominator contains no prime factors other than \(2\) and \(5\).

Factor the denominator:

\[
1400 = 14\cdot 100 = (2\cdot 7)(2^{2}\cdot 5^{2}) = 2^{3}\cdot 5^{2}\cdot 7 .
\]

The only ``bad'' prime factor is \(7\).   To eliminate this factor we must cancel it with a factor of \(7\) in the numerator. Hence \(n\) must be divisible by \(7\).

If \(n = 7k\) (with \(k\) an integer), then

\[
\frac{n}{1400}= \frac{7k}{7\cdot 2^{3}\cdot 5^{2}} = \frac{k}{2^{3}\cdot 5^{2}},
\]

and the remaining denominator consists solely of powers of \(2\) and \(5\), guaranteeing a terminating decimal. No further condition on \(k\) is needed.

Thus the problem reduces to counting the multiples of \(7\) between \(1\) and \(1000\):

\[
\left\lfloor\frac{1000}{7}\right\rfloor = 142.
\]

Therefore, there are

\[
\boxed{142}
\]

integer values of \(n\) in the range \(1\le n\le 1000\) for which \(\dfrac{n}{1400}\) terminates.
\end{quote}

\paragraph{Style 5 (default mix).}
2 teacher traces were assigned to this cluster; we show all of them.

\smallskip\noindent\textit{Teacher:} \texttt{qwen3.5-27b}

\begin{quote}
\small
Here's a thinking process that leads to the solution:

1.  \textbf{Analyze the Problem:}     *   We are looking for the number of integers $n$ such that $1 \le n \le 1000$.     *   The condition is that the decimal representation of the fraction $\frac{n}{1400}$ must terminate.

2.  \textbf{Recall the Condition for Terminating Decimals:}     *   A fraction $\frac{a}{b}$ (where $\gcd(a, b) = 1$) has a terminating decimal representation if and only if the prime factorization of the denominator $b$ contains *only* the prime factors 2 and 5.     *   If the fraction is not in simplest form, say $\frac{a}{b}$, we must first reduce it to lowest terms $\frac{a'}{b'}$. Then the condition is that $b'$ must be of the form $2^x 5^y$ for non-negative integers $x, y$.

3.  \textbf{Apply to the Specific Fraction:}     *   The fraction is $\frac{n}{1400}$.     *   Let's find the prime factorization of the denominator 1400.         *   $1400 = 14 \times 100$         *   $14 = 2 \times 7$         *   $100 = 10^2 = (2 \times 5)^2 = 2^2 \times 5^2$         *   So, $1400 = 2 \times 7 \times 2^2 \times 5^2 = 2^3 \times 5^2 \times 7^1$.

4.  \textbf{Simplify the Condition:}     *   Let the reduced fraction be $\frac{n'}{1400'}$.     *   We need the denominator $1400'$ (after canceling common factors with $n$) to have no prime factors other than 2 and 5.     *   The prime factors of 1400 are 2, 5, and 7.     *   The factors 2 and 5 are allowed. The factor 7 is *not* allowed.     *   Therefore, to make the decimal terminate, the factor 7 in the denominator must be canceled out by a factor of 7 in the numerator $n$.     *   This means $n$ must be divisible by 7.

5.  \textbf{Formulate the Counting Problem:}     *   We need to find the number of integers $n$ in the range $[1, 1000]$ such that $n$ is a multiple of 7.     *   Wait, is that sufficient?         *   If $n$ is a multiple of 7, let $n = 7k$.         *   Then $\frac{n}{1400} = \frac{7k}{1400} = \frac{k}{200}$.         *   The denominator is 200. $200 = 2^3 \times 5^2$.         *   The prime factors are only 2 and 5.         *   So, any multiple of 7 works.     *   Is it necessary?         *   Suppose $\frac{n}{1400}$ terminates.         *   Then the denominator in simplest form must be of the form $2^x 5^y$.         *   The original denominator is $2^3 \cdot 5^2 \cdot 7$.         *   For the 7 to disappear from the denominator, it must be a factor of the numerator $n$.         *   So yes, $n$ must be divisible by 7.

6.  \textbf{Calculate the Count:}     *   We need to count multiples of 7 between 1 and 1000 inclusive.     *   The multiples are $7 \times 1, 7 \times 2, \dots, 7 \times k$.     *   We need $7k \le 1000$.     *   $k \le \frac{1000}{7}$.     *   Let's perform the division:         *   $1000 \div 7 = 142.857\dots$         *   $140 \times 7 = 980$.         *   $142 \times 7 = 994$.         *   $143 \times 7 = 1001$.     *   So, the largest integer $k$ is 142.     *   Since $n$ starts at 1, $k$ starts at 1.     *   Therefore, there are 142 such values.

7.  \textbf{Double Check:}     *   Let $n = 7$. $\frac{7}{1400} = \frac{1}{200} = 0.005$. Terminates.     *   Let $n = 1$. $\frac{1}{1400} = \frac{1}{2^3 \cdot 5^2 \cdot 7}$. Denominator has factor 7. Repeats.     *   Let $n = 14$. $\frac{14}{1400} = \frac{1}{100} = 0.01$. Terminates.     *   The logic holds: we need to eliminate the factor 7 from the denominator.     *   The denominator is $1400 = 200 \times 7$.     *   $\frac{n}{1400} = \frac{n}{200 \times 7}$.     *   For this to terminate, the simplified denominator must not have any prime factors other than 2 and 5.     *   The prime factors of $1400$ are $\{2, 5, 7\}$.     *   The prime factors of the simplified denominator are $\{2, 5, 7\} \setminus \{p : p \text{ divides } \gcd(n, 1400)\}$.     *   For the set to be a subset of $\{2, 5\}$, the factor 7 must be removed.     *   This implies $7$ must divide $\gcd(n, 1400)$, which implies $7$ must divide $n$.     *   So $n$ must be a multiple of 7.     *   Range is $1 \le n \le 1000$.     *   Count is $\lfloor \frac{1000}{7} \rfloor$.     *   $1000 = 700 + 280 + 20$.     *   $1000 = 700 + 280 + 14 + 6$.     *   $1000 = 7(100 + 40 + 2) + 6$.     *   $1000 = 7(142) + 6$.     *   Quotient is 142.

8.  \textbf{Structure the Answer:}     *   State the condition for a terminating decimal.     *   Prime factorize the denominator 1400.     *   Determine the condition on $n$ (must be divisible by 7).     *   Calculate the number of multiples of 7 in the range [1, 1000].     *   State the final answer.

9.  \textbf{Refinement:}     *   Is there any edge case?     *   What if $n$ shares factors with 200? That doesn't matter, it just changes the power of 2 or 5 in the denominator, which is allowed.     *   Example: $n = 1400$. $\frac{1400}{1400} = 1$. Terminates. 1400 is a multiple of 7. But $n$ is restricted to 1000.     *   Example: $n = 700$. $\frac{700}{1400} = \frac{1}{2}$. Terminates.     *   Example: $n = 14$. $\frac{14}{1400} = \frac{1}{100}$. Terminates.     *   Example: $n = 2$. $\frac{2}{1400} = \frac{1}{700}$. Denominator has factor 7. Does not terminate.     *   The logic is solid.

10. \textbf{Final Calculation:}     *   $\lfloor 1000/7 \rfloor = 142$.

*   Wait, let's re-read the question carefully. "integer values of $n$ between 1 and 1000 inclusive". Yes.     *   "decimal representation ... terminate". Yes.

*
\end{quote}

\smallskip\noindent\textit{Teacher:} \texttt{qwen3-32b}

\begin{quote}
\small
<think> Okay, so I need to figure out how many integers n between 1 and 1000 inclusive make the decimal representation of n/1400 terminate. Hmm, terminating decimals... I remember that a fraction will have a terminating decimal if the denominator, after simplifying the fraction to its lowest terms, has no prime factors other than 2 and 5. Is that right? Let me confirm. Yeah, I think that's the rule. So if you can write the denominator as 2\^{}a * 5\^{}b where a and b are non-negative integers, then the decimal terminates. Otherwise, it repeats.

So first, I need to take the fraction n/1400 and simplify it. The key is that after simplifying, the denominator should only have 2s and 5s in its prime factorization. Let me recall that 1400 can be factored into primes. Let me factor 1400.

Starting with dividing by 2: 1400 ÷ 2 = 700. Again by 2: 700 ÷ 2 = 350. Again by 2: 350 ÷ 2 = 175. Now 175 is not divisible by 2. Let me check 5: 175 ÷ 5 = 35. Then 35 ÷ 5 = 7. And 7 is prime. So putting it all together, 1400 factors into 2\^{}3 * 5\^{}2 * 7\^{}1. So the prime factors are 2, 5, and 7.

Therefore, if we have n/1400, to have a terminating decimal, when we reduce the fraction, the denominator must not have any prime factors other than 2 and 5. Which means that any factor of 7 in the denominator must be canceled out by a factor of 7 in the numerator n. Therefore, n must be a multiple of 7. Is that correct? Let me think. If n shares a common factor with 1400, specifically the 7, then when we reduce the fraction, the 7 in the denominator will be canceled.

So for n/1400 to reduce to something with denominator only 2s and 5s, n must be divisible by 7. Let me check. Suppose n is divisible by 7. Then n = 7k for some integer k. Then n/1400 = 7k/(1400) = k/(200). Because 1400 divided by 7 is 200. And 200 factors into 2\^{}3 * 5\^{}2. Which only has 2s and 5s. So yes, if n is divisible by 7, then the reduced denominator is 200, so decimal terminates.

Conversely, if n is not divisible by 7, then when you reduce n/1400, there will still be a 7 in the denominator, making it non-terminating. Therefore, the decimal terminates if and only if n is divisible by 7.

Therefore, the problem reduces to finding the number of integers between 1 and 1000 inclusive that are divisible by 7. But wait, hold on. Is there a catch here? Because maybe when you reduce the fraction, even if n is divisible by 7, maybe there are other prime factors introduced? Wait, no. Because the original denominator is 2\^{}3 * 5\^{}2 * 7. If n is divisible by 7, then when you divide numerator and denominator by 7, you get denominator 2\^{}3 * 5\^{}2. Which is only 2s and 5s. So regardless of what else n is, as long as it cancels out the 7, the rest of the denominator is already okay.

But hold on, what if n has other common factors with the original denominator? For example, suppose n is divisible by 7 and also by 2 or 5. Then when you reduce the fraction, you would reduce the powers of 2 and 5 in the denominator. But even if you do that, the denominator still only has 2s and 5s. So even if n has common factors with 2 or 5, it doesn't affect the termination. Because the presence of 2s and 5s in the denominator is allowed. So the only problem is the 7.

Therefore, my conclusion is that n must be divisible by 7. Therefore, the number of such n between 1 and 1000 inclusive is equal to the number of multiples of 7 in that range.

But let me check with an example. Let's take n = 7. Then 7/1400 = 1/200 = 0.005, which terminates. Good. How about n = 14? 14/1400 = 1/100 = 0.01, terminates. n = 21: 21/1400 = 3/200 = 0.015, terminates. Okay. What if n is 7* something that also has 2s or 5s? Let's say n=14, which is 2*7. Then 14/1400 = 1/100. Which is 0.01. Still terminates. If n=35, which is 5*7. 35/1400 = 1/40 = 0.025. Terminates. If n=70, which is 2*5*7. 70/1400 = 1/20 = 0.05. Terminates. So even though n has factors of 2 and 5, as long as it cancels the 7, the decimal still terminates. Therefore, my conclusion is correct: n must be divisible by 7.

Therefore, the answer is the number of multiples of 7 between 1 and 1000 inclusive.

Now, how do we compute that? Well, the formula is floor(1000/7). But let me verify. The number of multiples of 7 between 1 and N inclusive is floor((N)/7). However, we need to check whether 1000 is divisible by 7. Let me compute 7*142 = 994. 7*143 = 1001. So 143*7 is 1001, which is over 1000. Therefore, the number of multiples of 7 between 1 and 1000 is 142. Wait, 7*142=994. Then 994 + 7 = 1001. So the last multiple of 7 less than or equal to 1000 is 994. Therefore, the number of multiples is 142. But wait, starting from 7*1=7, up to 7*142=994. So yes, 142 numbers. But let me check with the formula: floor(1000/7). Let me compute 1000 divided by 7.

7*142 = 994. 1000 - 994 = 6. So 1000/7 is 142 + 6/7, so floor(1000/7) is 142. Therefore, the number of multiples is 142.

But wait, hold on. The question says between 1 and 1000 inclusive. So n starts at 1. The first multiple of 7 is 7*1=7. So yes, so 142 numbers. Therefore, the answer is 142?

But let me check with another approach. Let me think if there's a possibility that even if n is divisible by 7, when you reduce n/1400, maybe the denominator could have other primes? Wait, suppose n is divisible by 7 but also by some other primes? Wait, no. Because if n has other primes, say 3, then when you reduce the fraction, would that prime remain in the denominator? Let me think.

Suppose n is 7*3 = 21. Then n/1400 = 21/1400 = 3/200. The denominator is 200, which is 2\^{}3 * 5\^{}2. So even though the numerator has a 3, which is a prime not in {2,5}, since it's in the numerator, it doesn't affect the denominator. Therefore, the denominator still only has 2s and 5s. Therefore, decimal terminates.

Another example: n=7*3* something. Let's take n=7*3*2=42. Then 42/1400 = 3*2/(200) = 6/200 = 3/100 = 0.03. Terminates.

So even if n has other prime factors besides 2 and 5 and 7, as long as it cancels the 7 in the denominator, the decimal will terminate. Therefore,
\end{quote}

\paragraph{Style 6 (long think-block).}
One teacher trace was assigned to this cluster.

\smallskip\noindent\textit{Teacher:} \texttt{phi-4-reasoning-plus}

\begin{quote}
\small
<think>We are asked: For how many integer values of n between 1 and 1000 inclusive does the decimal representation of n/1400 terminate?

We need to consider fraction n/1400, n is integer between 1 and 1000. We need to find number of n for which the fraction's decimal representation terminates. A rational fraction terminates if and only if after reduction, the denominator has only factors 2 and 5. Alternatively, we can say: fraction n/1400 terminates if and only if n is divisible by all prime factors of denominator aside from 2 and 5. Let's check factorization of 1400: 1400 = 14*100 = (2*7)* (2\^{}2*5\^{}2)= 2\^{}3 * 5\^{}2 *7. So the fraction is n / (2\^{}3 * 5\^{}2 * 7). For fraction to be terminating, after reducing, the denominator must be of the form 2\^{}a * 5\^{}b only. That means that the factor of 7 in denominator must be canceled out by factor in numerator. So n must be divisible by 7. But wait, is that sufficient? Let's check: if n is divisible by 7, then factor of 7 cancels out, leaving denominator of form 2\^{}3 * 5\^{}2 which is terminating. But is it necessary? Yes, if n is not divisible by 7, then the factor 7 remains in denominator and the fraction will be of the form something/7, which does not terminate (since 7 is not of the form 2\^{}a 5\^{}b). So the necessary and sufficient condition: n must be divisible by 7.

But wait, check: Actually, it's not exactly necessary that n must be divisible by 7. Wait, check: if n is not divisible by 7, then the denominator after reduction remains with factor 7, so it's not terminating. So indeed, n must be divisible by 7. So the answer: count multiples of 7 in the interval [1,1000]. How many multiples of 7 are there between 1 and 1000? The smallest multiple is 7, largest multiple is floor(1000/7)*7. floor(1000/7) = 142 because 7*142=994, 7*143=1001 >1000. So count = 142. But wait, double-check: Actually, is that the complete answer? Let's check: n/1400. But wait, is that sufficient? Let's test: n=7. Then fraction equals 7/1400 = 1/200. That terminates (1/200 = 0.005). Also n=14, then fraction equals 14/1400 = 1/100, which terminates. So yes.

Thus answer: 142.

Wait, but check: n=0? But n must be between 1 and 1000 inclusive. So yes.

Double-check: If n is divisible by 7, then n=7k. Then fraction equals (7k)/1400 = k/200, and since 200=2\^{}3*5\^{}2, so that fraction terminates. So answer: 142.

So answer: 142.

But wait, we must check if there's any possibility that n might have factors that cancel some factors of 2 or 5? But that's not necessary. Actually, wait, let me re-read the condition for a fraction a/b in simplest form to have terminating decimal representation. It is that b's prime factors can only be 2 and 5. That is exactly what we are doing. So yes.

Thus answer: 142.

Double-check: The number of multiples of 7 between 1 and 1000 equals floor(1000/7) = 142.

Thus answer: 142.

I'll now produce answer: 142.

I'll now produce answer in plain text.

I'll produce answer in plain text with explanation: "The fraction n/1400 terminates if and only if the reduced denominator has no prime factors other than 2 and 5. Since 1400=2\^{}3 * 5\^{}2 *7, the factor 7 must be cancelled by n. Therefore, n must be divisible by 7. The number of multiples of 7 between 1 and 1000 is floor(1000/7)=142. Therefore, the answer is 142."

I'll produce answer with explanation.

I'll now produce answer.</think>To determine when the decimal representation of n/1400 terminates, we start by factoring the denominator:

1400 = 14 \(\times\) 100 = (2 \(\times\) 7) \(\times\) (2² \(\times\) 5²) = 2³ \(\times\) 5² \(\times\) 7.

A fraction in lowest terms has a terminating decimal if and only if its denominator has no prime factors other than 2 and 5. In the fraction n/1400, the factor 7 must be canceled out by n. This means n must be divisible by 7.

Now, we need to count how many integers n between 1 and 1000 (inclusive) are divisible by 7. The smallest multiple of 7 in this range is 7, and the largest is found by dividing 1000 by 7:

1000 ÷ 7 = 142.857, so the largest integer multiple is 7 \(\times\) 142 = 994.

Thus, there are 142 multiples of 7 between 1 and 1000.

Therefore, there are 142 integer values of n between 1 and 1000 for which n/1400 has a terminating decimal representation.
\end{quote}

\subsection{LLM based style discovery}
\label{app:llmstyledicovery}
As a human-readable check on the autoencoder taxonomy, we asked a long-context LLM
to invent writing-style clusters from unlabeled SCAS teacher writeups.
The judge was Cursor Grok~4.6
(\texttt{cursor-grok-4.6-high-fast}~\citet{grok46modelcard2026}), run as a fresh Cursor agent with no
access to teacher IDs, question text, or GMM labels.
We report the primary $180$-trace protocol below (a $45$-trace pilot on five
style-balanced questions gave the same seven-style inventory).

\paragraph{Sample.}
$180$ traces: $20$ per teacher, distinct questions (excluding the five-question
pilot), IDs shuffled as \texttt{T001}--\texttt{T180}.
Long writeups were truncated in the middle ($200$ opening words + $80$ closing
words); the prompt told the model to judge style from opening and closing form.

\paragraph{Prompt (instructions only).}
The writeups then followed as \texttt{===== Txxx =====} blocks.
\begin{quote}\small\ttfamily
You are given unlabeled math solution writeups. Do not assume how many styles
exist or who wrote them. Some writeups are truncated in the middle; judge style
from the opening and closing form.\\[0.4em]
Task:\\
1. Infer writing-style clusters from formatting, voice, structure, and surface
form --- not from the math content or the final numeric answer.\\
2. Decide the number of styles yourself. Name each style with a short label and
2--4 sentence description of the distinctive cues.\\
3. Assign every ID to exactly one style.\\
4. If two writeups solve different problems but share the same manner of writing,
they belong together.\\[0.4em]
Return ONLY:\\
- A list of styles: label, description, member IDs\\
- A complete ID to style assignment\\
- A brief note on any IDs that were hard to place\\[0.4em]
Do not mention teachers, models, clusters, autoencoders, or GMM.
\end{quote}

\paragraph{Agreement.}
On this sample, chance-corrected agreement is
LLM--teacher $\mathrm{ARI}{=}0.70$,
LLM--autoencoder $\mathrm{ARI}{=}0.07$,
autoencoder--teacher $\mathrm{ARI}{=}0.06$.
The LLM inventory tracks teacher surface fingerprints much more than the
$K{=}6$ GMM bins.

\begin{table}[t]
\centering
\caption{Blind LLM styles on the $180$-trace sample ($20$ writeups per teacher)
vs.\ the oracle teacher that produced each writeup.
Most styles recover a single teacher at $18$--$20/20$; the markdown-tutor bin
merges gpt-5 and olmo; the instructional walkthrough merges gemma and qwen2.5.}
\label{tab:llm-blind-style-teachers}
\small
\setlength{\tabcolsep}{4pt}
\begin{tabular}{llc}
\toprule
LLM style & $n$ & Teacher mix (of 20 per teacher) \\
\midrule
Exploratory first-person monologue & 20 & qwen3 $20/20$ \\
Telegraphic ``We are asked'' think-block & 20 & phi-4 $20/20$ \\
Structured request-analysis outline & 18 & qwen3.5 $18/20$ \\
H2-numbered recipe with stock closer & 20 & llama $20/20$ \\
Restating markdown tutor & 34 & gpt-5 $18/20$, olmo $16/20$ \\
Compact contest exposition & 25 & gpt-oss $20/20$; also gpt-5\,2, qwen3.5\,2, qwen2.5\,1 \\
Instructional ``To find\ldots'' walkthrough & 43 & gemma $20/20$, qwen2.5 $19/20$; also olmo\,4 \\
\bottomrule
\end{tabular}
\end{table}

\begin{table}[t]
\centering
\caption{Same $180$-trace sample, transposed: for each teacher, how many of its
$20$ writeups the LLM assigned to each invented style.
Pure diagonals (e.g.\ llama$\rightarrow$H2 recipe, gpt-oss$\rightarrow$contest)
are the recoverable fingerprints; olmo and gpt-5 share the markdown-tutor bin.}
\label{tab:llm-blind-teacher-styles}
\scriptsize
\setlength{\tabcolsep}{3pt}
\begin{tabular}{lccccccc}
\toprule
Teacher & mono. & think & outline & H2 & md tutor & contest & walk. \\
\midrule
gemma-4-31b-it & 0 & 0 & 0 & 0 & 0 & 0 & 20 \\
gpt-5-chat & 0 & 0 & 0 & 0 & 18 & 2 & 0 \\
gpt-oss-120b & 0 & 0 & 0 & 0 & 0 & 20 & 0 \\
llama-3.3-70b-instruct & 0 & 0 & 0 & 20 & 0 & 0 & 0 \\
olmo-3.1-32b-instruct & 0 & 0 & 0 & 0 & 16 & 0 & 4 \\
phi-4-reasoning-plus & 0 & 20 & 0 & 0 & 0 & 0 & 0 \\
qwen2.5-72b-instruct & 0 & 0 & 0 & 0 & 0 & 1 & 19 \\
qwen3-32b & 20 & 0 & 0 & 0 & 0 & 0 & 0 \\
qwen3.5-27b & 0 & 0 & 18 & 0 & 0 & 2 & 0 \\
\bottomrule
\end{tabular}\\[0.3em]
{\footnotesize Column keys match \cref{tab:llm-blind-style-teachers}:
mono.\ $=$ exploratory monologue;
think $=$ telegraphic think-block;
outline $=$ request-analysis;
H2 $=$ H2-numbered recipe;
md tutor $=$ restating markdown;
contest $=$ compact contest;
walk.\ $=$ instructional walkthrough.}
\end{table}

\paragraph{Model output: seven styles.}
Verbatim from the Grok~4.6 response (member counts only; full ID lists omitted).

\begin{enumerate}
\item \textbf{Exploratory first-person monologue} ($n{=}20$).
Opens in a talking-to-yourself voice
(\verb|Okay, so I need to...|, \verb|Hmm|, \verb|Let me think|, \verb|Alright|).
Often wrapped in a \verb|<think>| block.
The writeup rambles, restates the problem, second-guesses, and narrates memory
of formulas before settling.
Closings, when present, are an afterthought rather than a fixed template.

\item \textbf{Telegraphic ``We are asked'' think-block} ($n{=}20$).
Opens \verb|<think>We are asked:| or \verb|<think>We are given:| with almost
no space after the tag.
Dense, repetitive inner speech
(\verb|We'll solve|, \verb|Thus answer:|,
\verb|I'll produce answer|~/~\verb|We'll produce answer|).
After the block (or after a truncation cut) a cleaner rewrite often appears.
Surface form is compressed and formulaic, not conversational.

\item \textbf{Structured request-analysis outline} ($n{=}18$).
Opens with a meta header such as
\verb|Here's a thinking process that leads to the solution|,
\verb|Here's my thought process|,
\verb|Thinking Process:|, or
\verb|The user wants to...|.
Numbered stages with nested bullets:
\textbf{Analyze the Request}, \textbf{Understand the Goal},
Given~/~Constraints~/~Strategy~/~Execution.
Treats the problem as a request to be dissected before computing.

\item \textbf{H2-numbered recipe with stock closer} ($n{=}20$).
Every piece starts at \verb|## Step 1:| (exactly two hashes) and walks through
\verb|## Step 2|, \verb|## Step 3|, \ldots\ in running prose with inline
\verb|$...$| math.
Always ends with the same closer:
\verb|The final answer is: $\boxed{...}$|.
No checkmarks, no \verb|<think>|, no ``Let's restate.''

\item \textbf{Restating markdown tutor} ($n{=}34$).
Friendly coach voice:
\verb|Let's restate the problem carefully|,
\verb|Let's carefully go step-by-step|,
\verb|Let's break down the problem|.
Heavy markdown: \verb|---| rules,
\verb|### Given|~/~\verb|### Step|, bold key phrases, and often a
leading checkmark before \textbf{Final Answer} and \verb|\boxed{}|.
Even when the emoji is missing, the restating/step-card layout is the same.

\item \textbf{Compact contest exposition} ($n{=}25$).
Starts in the middle of the math:
\verb|Let ...|, \verb|The ...|, \verb|First ...|, or a display
\verb|\[|.
Tight declarative sentences,
\verb|aligned|~/~\verb|\qquad|~/~\verb|\Longrightarrow|,
lists of values, then a lone \verb|\boxed{}|.
Little student-facing scaffolding and almost no step theater.

\item \textbf{Instructional ``To find\ldots'' walkthrough} ($n{=}43$).
Opens \verb|To find|~/~\verb|To solve|~/~\verb|To determine|~/~\verb|To compute|
(or a close \verb|We are given|~/~\verb|Let $x$ be| plus ``we can'').
Numbered \verb|1. **Bold action:**| lists or sequential
``we start~/~next~/~thus'' prose.
Homework-helper voice; closer is \verb|\boxed{}|,
\verb|**Answer:**|, or a plain concluding sentence.
Distinct from style~5 by the task-announcing lead-in and the absence of
restating cards~/~checkmarks.
\end{enumerate}

\paragraph{Hard-to-place notes (model).}
The judge flagged borderline openings that mix tutor restating with first-person
self-correction, \verb|To determine...| leads that continue as markdown cards,
short we-voice walkthroughs that resemble contest setups, and
\verb|## Step| headings that close like style~5 rather than the stock
``The final answer is:'' recipe.
Assignments followed the dominant surface cues (opening voice, headers, closer).

\section{Importance-Weighted SFT Variants}
\label{app:importance-weighting}

We compare three ways of reweighting the same set of $N=111{,}834$ training traces. In all cases, the underlying examples are unchanged; only their contribution to the token-level negative log-likelihood differs. Importance weights are mean-normalized to one before training, and we use $\varepsilon=0.01$ for smoothing.

Let $C=6$ denote the number of discovered styles. Question $x_i$ has $K_i$ teacher traces with style assignments
$c_i^1,\ldots,c_i^{K_i}$. We define
\begin{equation}
n_{i,c}
=
\sum_{k=1}^{K_i}
\mathbb{I}(c_i^k=c),
\end{equation}
and
\begin{equation}
S_i
=
\{c:n_{i,c}>0\},
\end{equation}
where $n_{i,c}$ is the number of traces for question $x_i$ assigned to style $c$, and $S_i$ is the set of styles actually observed for that question.

For the two question-conditional variants, we use the same smoothed proposal distribution. Before normalization across styles,
\begin{equation}
q_{\mathrm{raw}}(c\mid x_i)
=
\frac{
\sum_{k=1}^{K_i}
\left(
\mathbb{I}(c_i^k=c)+\varepsilon
\right)}
{K_i(1+\varepsilon)}.
\end{equation}
Since
\begin{equation}
\sum_{c'=1}^{C}
q_{\mathrm{raw}}(c'\mid x_i)
=
\frac{1+C\varepsilon}{1+\varepsilon},
\end{equation}
renormalizing over the $C$ styles gives
\begin{equation}
q(c\mid x_i)
=
\frac{n_{i,c}+K_i\varepsilon}
     {K_i(1+C\varepsilon)}.
\end{equation}

Each training trace receives an importance weight
\begin{equation}
w
=
\frac{p}{q},
\end{equation}
where $p$ is the target style distribution for the corresponding arm. Before optimization, we mean-normalize the weights,
\begin{equation}
w_j
\leftarrow
\frac{w_j}{\bar{w}},
\end{equation}
and optimize the weighted token-level negative log-likelihood
\begin{equation}
\mathcal{L}
=
\frac{1}{|B|}
\sum_{j\in B}
w_j\,
\ell\!\left(
x_j,(c_j,r_j),y_j
\right).
\end{equation}

\subsection{Per-question uniform target over all styles: IS-rebal}
\label{app:is-rebal}

In \textsc{IS-rebal}, the target distribution is uniform over all $C$ discovered styles for every question:
\begin{equation}
p_{\mathrm{rebal}}(c\mid x_i)
=
\frac{1}{C}.
\end{equation}
The corresponding importance weight is
\begin{equation}
w_{\mathrm{rebal}}(c\mid x_i)
=
\frac{1/C}
     {q(c\mid x_i)}.
\end{equation}

This can be viewed as approximating the expectation under a per-question uniform style distribution:
\begin{equation}
\mathbb{E}_{c\sim p_{\mathrm{rebal}}(c\mid x_i)}
\left[\ell\right]
=
\mathbb{E}_{c\sim q(c\mid x_i)}
\left[
\frac{p_{\mathrm{rebal}}(c\mid x_i)}
     {q(c\mid x_i)}
\ell
\right].
\end{equation}

Because the finite training set contains no trace for styles absent from $S_i$, the Monte Carlo estimator only receives contributions from observed styles. The target nevertheless assigns mass to all $C$ styles, while smoothing ensures $q(c\mid x_i)>0$ for every style. Consequently, the raw target mass represented by the available traces for question $x_i$ is
\begin{equation}
\frac{|S_i|}{C}.
\end{equation}

Thus, \textsc{IS-rebal} is the most aggressive question-conditional balancing scheme: it treats all six styles as equally desirable for every question, even when only a subset of them is represented in the teacher traces.

\subsection{Per-question uniform target over observed styles: IS-obs}
\label{app:is-obs}

In \textsc{IS-obs}, we instead define the target distribution only over styles that actually occur among the teacher traces for question $x_i$:
\begin{equation}
p_{\mathrm{obs}}(c\mid x_i)
=
\frac{\mathbb{I}(c\in S_i)}
     {|S_i|}.
\end{equation}
The importance weight is therefore
\begin{equation}
w_{\mathrm{obs}}(c\mid x_i)
=
\frac{
\mathbb{I}(c\in S_i)/|S_i|
}
{q(c\mid x_i)}.
\end{equation}

Equivalently,
\begin{equation}
\mathbb{E}_{c\sim p_{\mathrm{obs}}(c\mid x_i)}
\left[\ell\right]
=
\mathbb{E}_{c\sim q(c\mid x_i)}
\left[
\frac{p_{\mathrm{obs}}(c\mid x_i)}
     {q(c\mid x_i)}
\ell
\right].
\end{equation}

This variant equalizes only the styles for which a trace is actually available. For example, if a question has eight traces in $s_5$ and one trace in $s_2$, then
\begin{equation}
S_i=\{s_2,s_5\},
\end{equation}
and the target assigns probability $1/2$ to each of these styles and zero to the remaining four. Thus, \textsc{IS-obs} rebalances within the observed support of each question without assigning target mass to unobserved styles.

\subsection{Dataset-wide uniform target: IS-global}
\label{app:is-global}

The third variant ignores question identity and reweights styles according to their global frequency in the training corpus. Let
\begin{equation}
N_c
=
\sum_{i,k}
\mathbb{I}(c_i^k=c),
\end{equation}
and
\begin{equation}
N
=
\sum_{c=1}^{C}N_c.
\end{equation}
We define the smoothed global proposal
\begin{equation}
q(c)
=
\frac{N_c+\varepsilon}
     {N+C\varepsilon},
\end{equation}
and use a uniform target over styles,
\begin{equation}
p_{\mathrm{global}}(c)
=
\frac{1}{C}.
\end{equation}
The resulting weight is
\begin{equation}
w_{\mathrm{global}}(c)
=
\frac{1/C}{q(c)}
\propto
\frac{1}{N_c+\varepsilon}.
\end{equation}

The corresponding importance-sampling identity is
\begin{equation}
\mathbb{E}_{c\sim p_{\mathrm{global}}(c)}
\left[\ell\right]
=
\mathbb{E}_{c\sim q(c)}
\left[
\frac{p_{\mathrm{global}}(c)}
     {q(c)}
\ell
\right].
\end{equation}

Unlike the two question-conditional variants, \textsc{IS-global} assigns one weight to each style for the entire corpus. Frequent styles such as $s_5$ are downweighted and rarer styles are upweighted, but the weighting does not depend on which styles are available for a particular question.

\subsection{Summary of weighting schemes}

\begin{figure}[htbp]
    \centering
    \includegraphics[width=\textwidth]{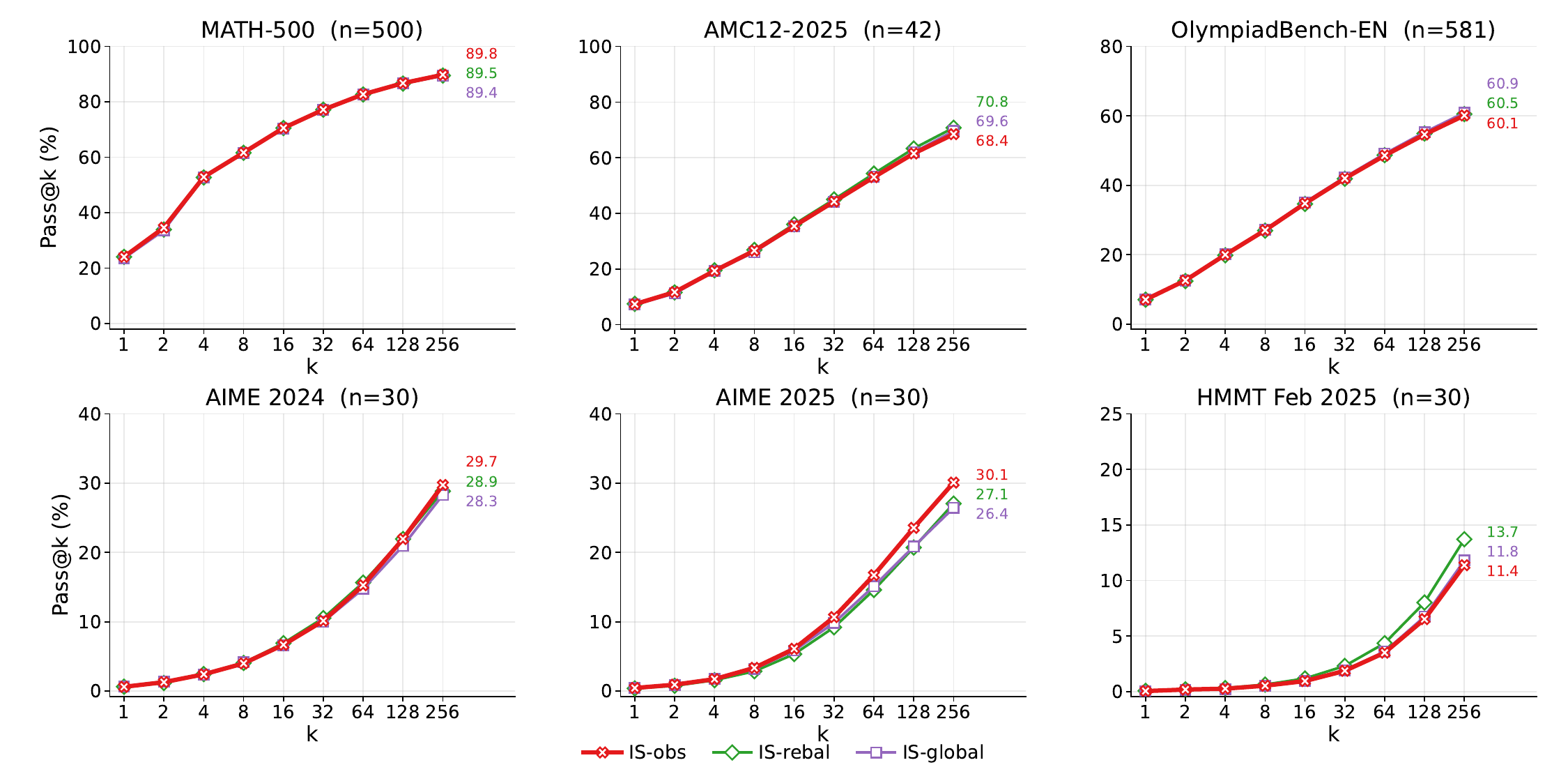}
    \caption{Qwen3 0.6B importance sampling variants}
    \label{fig:0p6bpassk_is}
\end{figure}

\begin{figure}[htbp]
    \centering
    \includegraphics[width=\textwidth]{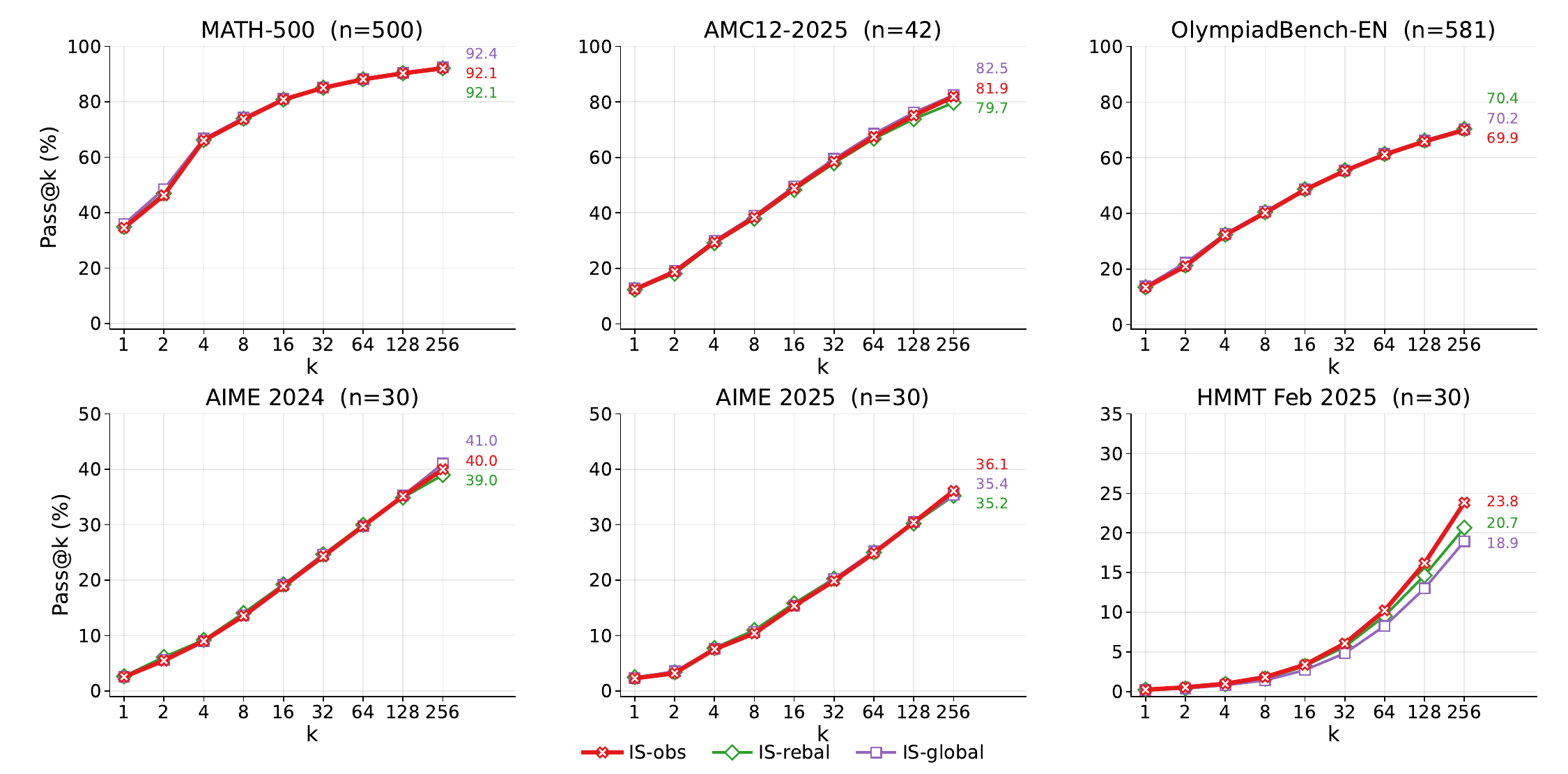}
    \caption{Qwen3 1.7B importance sampling variants}
    \label{fig:1p7bpassk_is}
\end{figure}

The three variants differ only in the target distribution used to define the importance ratio:
\begin{equation}
\begin{aligned}
\textsc{IS-rebal}:&\quad
p(c\mid x_i)=\mathrm{Unif}(\{1,\ldots,C\}),\\
\textsc{IS-obs}:&\quad
p(c\mid x_i)=\mathrm{Unif}(S_i),\\
\textsc{IS-global}:&\quad
p(c)=\mathrm{Unif}(\{1,\ldots,C\}).
\end{aligned}
\end{equation}
All three retain the full set of training traces and differ only in how those traces are weighted during SFT. We report the downstream results of these three weighting schemes for Qwen3 0.6B~(Figure~\ref{fig:0p6bpassk_is}), 1.7B~(Figure~\ref{fig:1p7bpassk_is}), and 4B~(Figure~\ref{fig:4bpassk_is}). We used IS-obs for the experiments in the main paper and the rest of experiments in the appendix.

\begin{figure}[htbp]
    \centering
    \includegraphics[width=\textwidth]{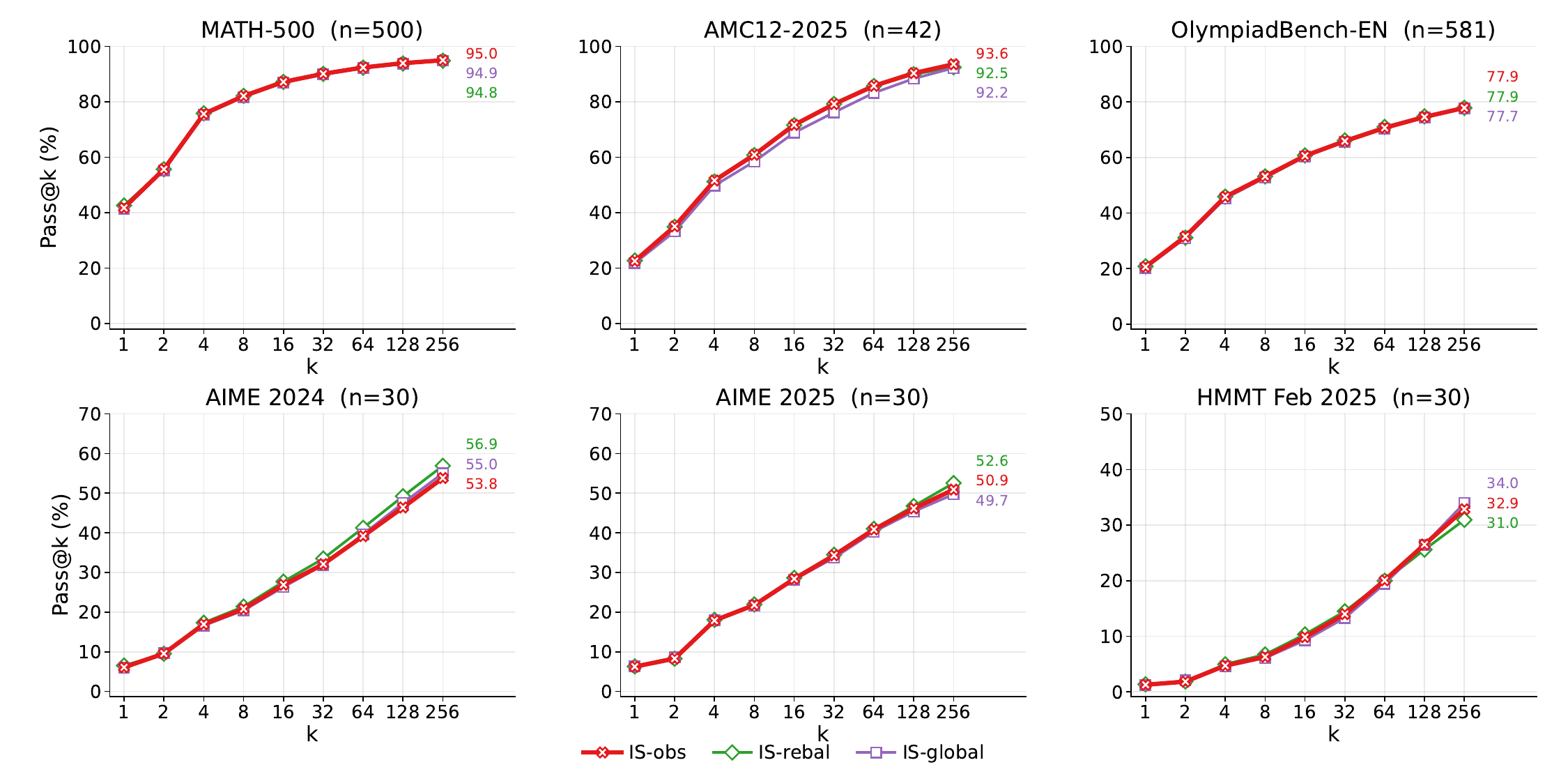}
    \caption{Qwen3 4B importance sampling variants}
    \label{fig:4bpassk_is}
\end{figure}

\section{Full results with Qwen3 1.7B and 4B}
\begin{figure}[htbp]
    \centering
    \includegraphics[width=\textwidth]{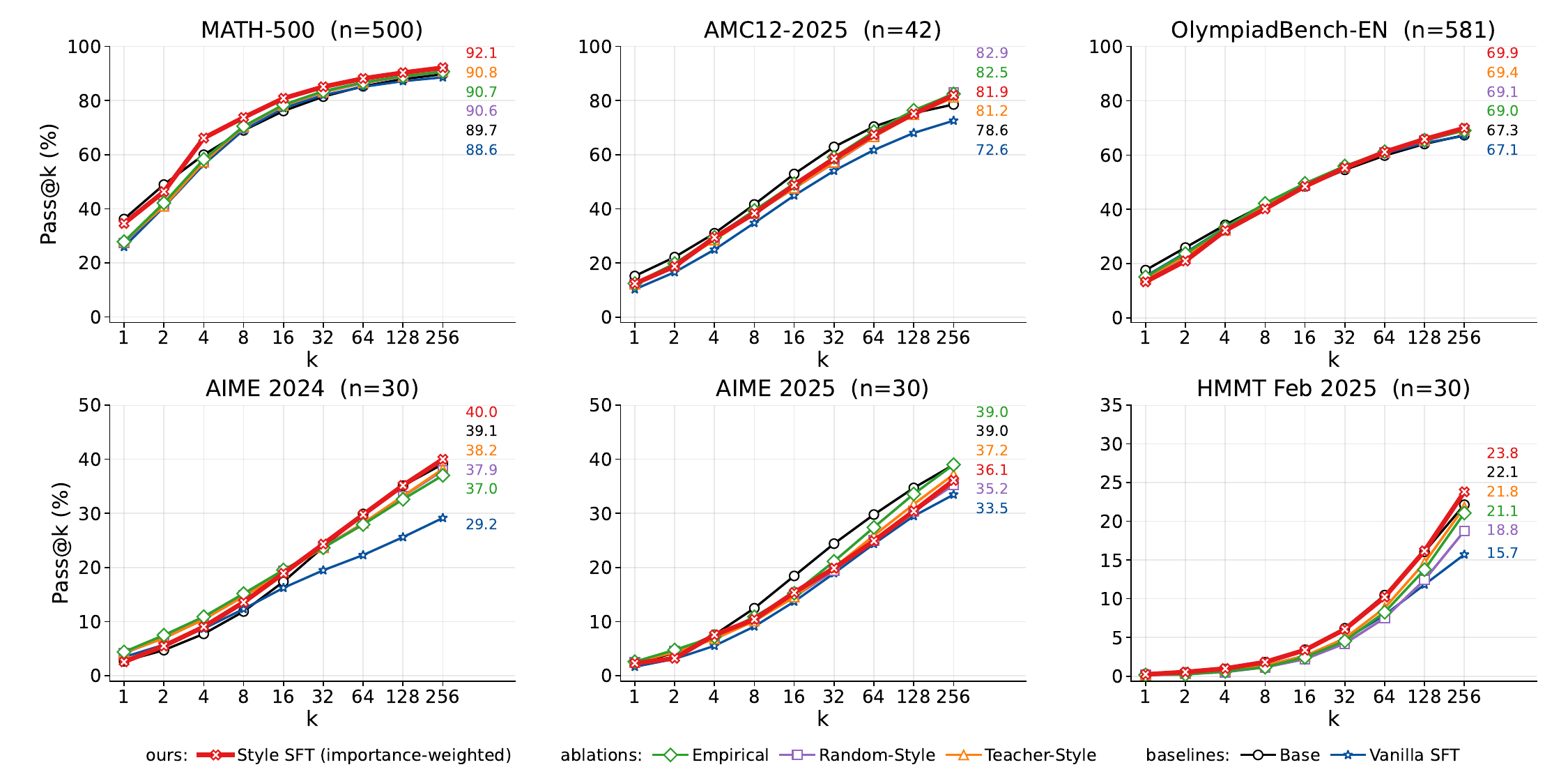}

    \caption{Pass@$k$ Qwen3 1.7B}
    \label{fig:1p7bpassk}
\end{figure}

\begin{figure}[htbp]
    \centering
    \includegraphics[width=\textwidth]{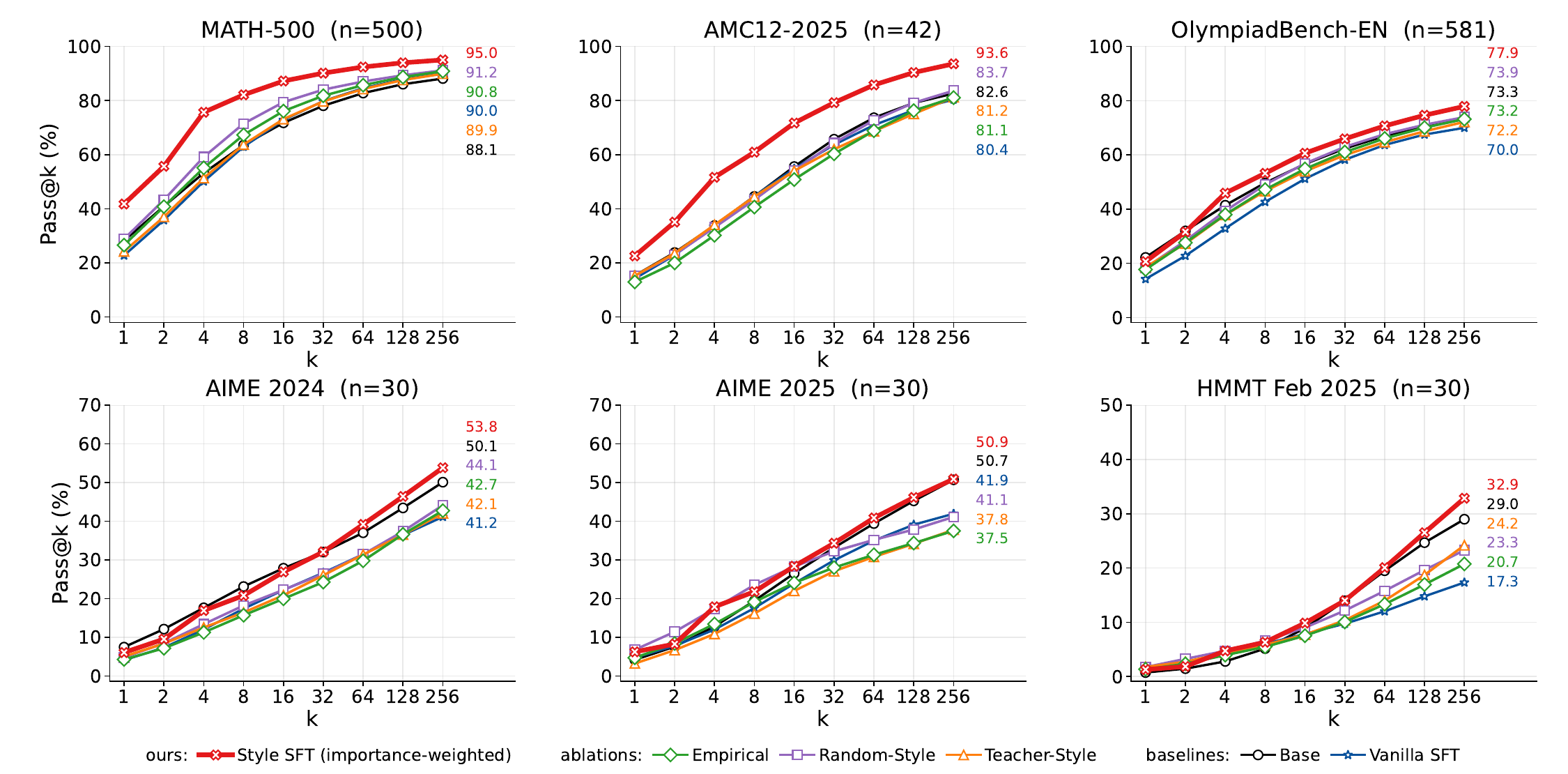}

    \caption{Pass@$k$ Qwen3 4B}
    \label{fig:4bpassk}
\end{figure}

We report the Pass@$k$ for Qwen3 1.7B in Figure~\ref{fig:1p7bpassk} and 4B  Figure~\ref{fig:4bpassk}.
\label{app:fullresults}

\newpage
\subsection{Style-Prefix Adherence}
\label{app:style-adherence}

We evaluate whether the discrete style prefixes learned during SFT induce generations that resemble the corresponding styles discovered in the teacher traces. This analysis distinguishes between a prefix acting as an arbitrary control token and a prefix that actually steers generation toward the associated teacher-style mode.

For each MATH-500 generation produced under a requested prefix
$\texttt{[style\_i]}$, we assign a realized style $\hat{s}$ using the same
handcrafted style feature space used to characterize the teacher clusters.
The feature representation consists of length-normalized structural features from Table~\ref{tab:style-profile-full}
together with log word count. We standardize generated traces using statistics
computed from the teacher corpus and assign each generation to its nearest
teacher-style centroid. We then compute the row-normalized confusion matrix
\begin{equation}
P\left(
\hat{s}=j
\mid
\texttt{[style\_i]}
\right),
\end{equation}
where rows correspond to the requested prefix and columns to the realized
teacher-style assignment. Perfect control would place all mass on the
diagonal, while a uniform assignment over the six styles would give diagonal
mass of $1/6=16.7\%$.

\begin{figure*}[t]
    \centering

    \begin{minipage}[t]{\textwidth}
        \centering
        \includegraphics[width=\linewidth]{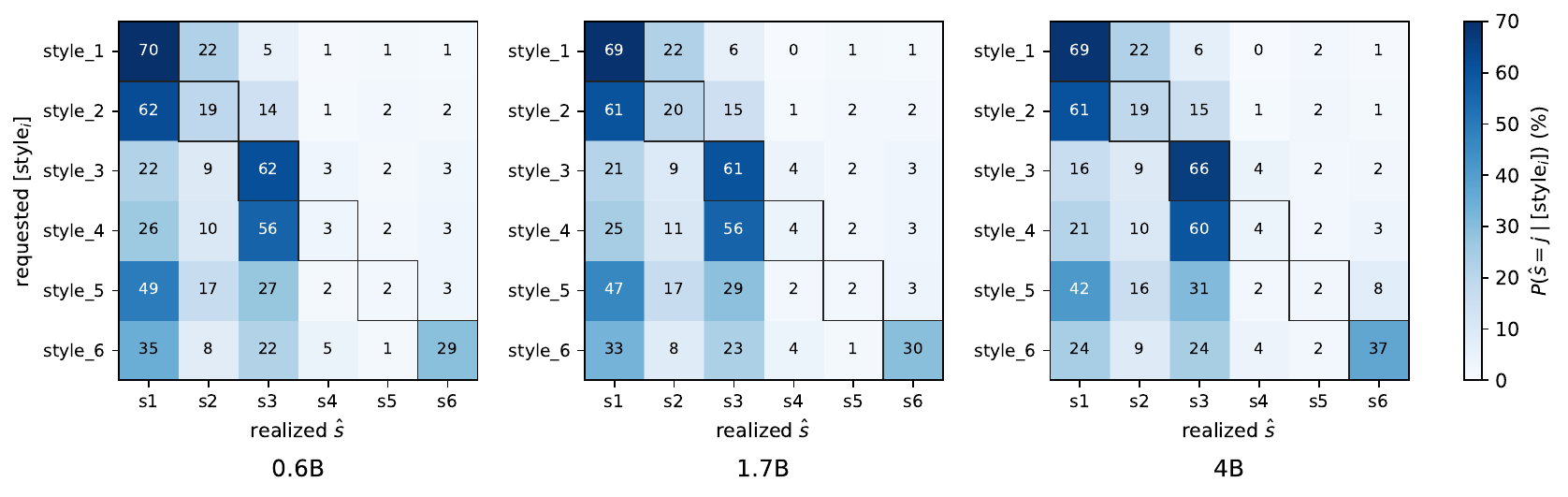}
        
        \small (a) \textsc{Style SFT (Importance-Weighted)}
    \end{minipage}
    \hfill
    \begin{minipage}[t]{\textwidth}
        \centering
        \includegraphics[width=\linewidth]{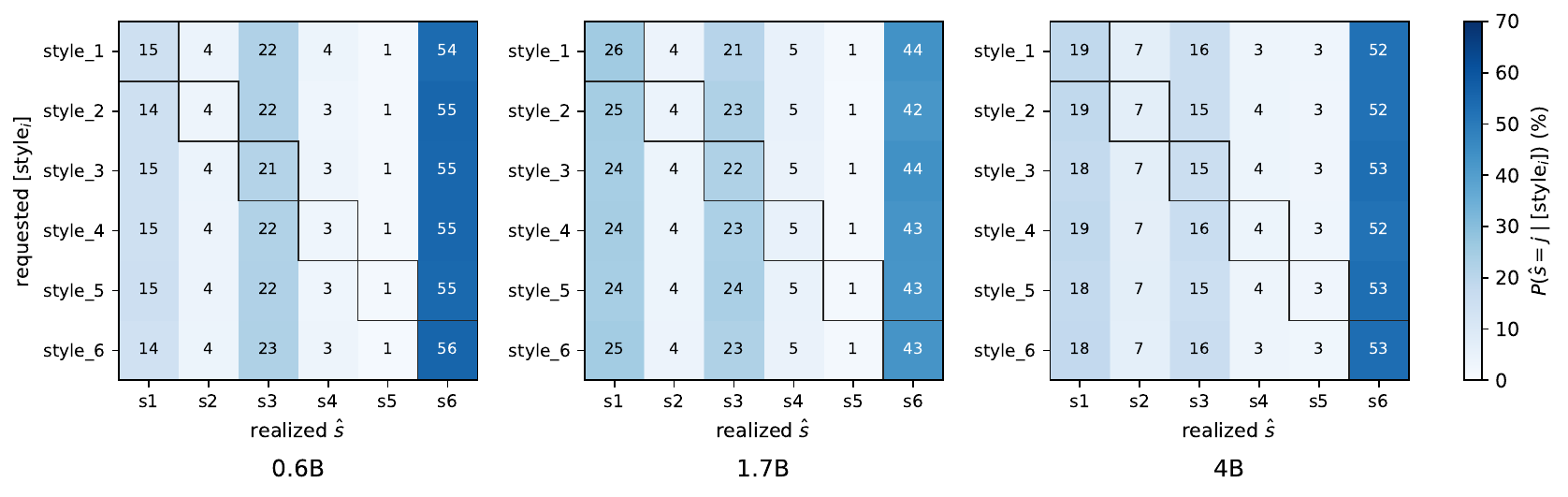}
        
        \small (b) \textsc{Style SFT (Empirical)}
    \end{minipage}

    \caption{
    Requested versus realized style on MATH-500.
    Rows denote the forced prefix $\texttt{[style\_i]}$ and columns denote the
    realized style $\hat{s}$ assigned by the nearest teacher-style centroid in
    the handcrafted style feature space.
    Values report
    $P(\hat{s}=j\mid\texttt{[style\_i]})$.
    Diagonal mass measures adherence to the requested teacher-style mode,
    while off-diagonal mass reveals which styles are conflated.
    (a) Importance-weighted Style SFT yields substantially greater style
    fidelity than (b) empirical Style SFT, although adherence remains uneven
    across modes.
    }
    \label{fig:style-adherence}
\end{figure*}

Figure~\ref{fig:style-adherence} compares the full importance-weighted model
with \textsc{Style SFT (Empirical)}. For
\textsc{Style SFT (Importance-Weighted)}, mean diagonal agreement is
$30.8\%$, $30.8\%$, and $32.9\%$ for the 0.6B, 1.7B, and 4B students,
respectively, approximately twice the chance level. Controllability is
strongly style-dependent. Requested $s_1$ and $s_3$ are realized reliably,
with roughly $60$--$70\%$ of generations assigned to the corresponding
teacher mode. Style $s_6$ shows moderate adherence, while $s_2$, $s_4$, and
$s_5$ are frequently assigned to neighboring modes, particularly $s_1$ and
$s_3$.

In contrast, \textsc{Style SFT (Empirical)} remains close to chance overall.
Thus, simply attaching style labels to the empirical training distribution
does not cause the student to reproduce the corresponding teacher-style
structure reliably. Importance-weighted training substantially strengthens
the association between the requested prefix and the realized style, although
the six modes are not perfectly separable in the student.

These results support a partial-controllability interpretation of the style
prefixes. The strongest prefixes recover identifiable teacher-style modes,
while others remain conflated. We therefore treat the prefixes as controls
associated with the discovered style structure rather than as perfect
realizations of six independent styles.

\newpage

\subsection{Examples of Style-Conditioned Generations}
\label{app:style-examples}

To qualitatively illustrate the discovered styles, we show generations from
the same MATH-500 problem under each of the six style prefixes. Holding the
question fixed makes differences in organization, verbosity, and formatting
easier to compare. These examples are generated by the Qwen3-0.6B student
trained with importance-weighted Style SFT.

\paragraph{Question.}
Bill walks $\frac{1}{2}$ mile south, then $\frac{3}{4}$ mile east, and
finally $\frac{1}{2}$ mile south. How many miles is he, in a direct line,
from his starting point? Express your answer as a decimal to the nearest
hundredth.

\paragraph{Gold answer.}
$\boxed{1.25}$

\medskip
\hrule
\medskip

\paragraph{$s_1$ --- short textbook.}
\begin{quote}
\small
\textbf{Step 1: Calculate the total distance Bill walks south.}

Bill walks $\frac{1}{2}$ mile south and then $\frac{1}{2}$ mile south.
So, the total distance he walks south is
$\frac{1}{2}+\frac{1}{2}=1$ mile.

\textbf{Step 2: Calculate the total distance Bill walks east.}

Bill walks $\frac{3}{4}$ mile east.

\textbf{Step 3: Apply the Pythagorean theorem.}

The direct distance is
\[
\sqrt{1^2+\left(\frac{3}{4}\right)^2}
=
\sqrt{\frac{25}{16}}
=
\frac{5}{4}
=
1.25.
\]

The final answer is: $\boxed{1.25}$.
\end{quote}

\medskip
\hrule
\medskip

\paragraph{$s_2$ --- short unmarked.}
\begin{quote}
\small
Bill walks $\frac{1}{2}$ mile south and then another $\frac{1}{2}$ mile
south, giving a net displacement of $1$ mile south. He also walks
$\frac{3}{4}$ mile east.

The distance from the starting point is therefore
\[
d
=
\sqrt{(-1)^2+\left(\frac{3}{4}\right)^2}
=
\sqrt{1+\frac{9}{16}}
=
\sqrt{\frac{25}{16}}
=
\frac{5}{4}
=
1.25.
\]

The final answer is: $\boxed{1.25}$.
\end{quote}

\medskip
\hrule
\medskip

\paragraph{$s_3$ --- contest markdown.}
\begin{quote}
\small
Let Bill start at the origin $(0,0)$.

\begin{enumerate}
    \item After walking $\frac{1}{2}$ mile south:
    \[
    \left(0,-\frac{1}{2}\right).
    \]

    \item After walking $\frac{3}{4}$ mile east:
    \[
    \left(\frac{3}{4},-\frac{1}{2}\right).
    \]

    \item After walking another $\frac{1}{2}$ mile south:
    \[
    \left(\frac{3}{4},-1\right).
    \]
\end{enumerate}

Hence the distance from the origin is
\[
\sqrt{\left(\frac{3}{4}\right)^2+(-1)^2}
=
\sqrt{\frac{25}{16}}
=
\frac{5}{4}
=
1.25.
\]

Thus,
\[
\boxed{1.25}.
\]
\end{quote}

\medskip
\hrule
\medskip

\paragraph{$s_4$ --- longer markdown.}
\begin{quote}
\small
Bill's net displacement is the sum of his movements:

\begin{itemize}
    \item South: $-\frac{1}{2}$ mile
    \item East: $\frac{3}{4}$ mile
    \item South: $-\frac{1}{2}$ mile
\end{itemize}

The total southward displacement is
\[
-\frac{1}{2}-\frac{1}{2}=-1,
\]
while the eastward displacement is $\frac{3}{4}$ mile.

The straight-line distance from the starting point is therefore
\[
\sqrt{(-1)^2+\left(\frac{3}{4}\right)^2}
=
\sqrt{1+\frac{9}{16}}
=
\sqrt{\frac{25}{16}}
=
\frac{5}{4}
=
1.25.
\]

\[
\boxed{1.25}
\]
\end{quote}

\medskip
\hrule
\medskip

\paragraph{$s_5$ --- default mix.}
\begin{quote}
\small
Let's break down Bill's journey step by step. Suppose he starts at the
origin $(0,0)$.

\textbf{First leg.}
He walks $\frac{1}{2}$ mile south, reaching
\[
\left(0,-\frac{1}{2}\right).
\]

\textbf{Second leg.}
He walks $\frac{3}{4}$ mile east, reaching
\[
\left(\frac{3}{4},-\frac{1}{2}\right).
\]

\textbf{Third leg.}
He walks another $\frac{1}{2}$ mile south, reaching
\[
\left(\frac{3}{4},-1\right).
\]

Using the distance formula,
\[
d
=
\sqrt{\left(\frac{3}{4}\right)^2+(-1)^2}
=
\sqrt{\frac{25}{16}}
=
\frac{5}{4}
=
1.25.
\]

\textbf{Final answer:}
\[
\boxed{1.25}
\]
\end{quote}

\medskip
\hrule
\medskip

\paragraph{$s_6$ --- long think-block.}
\begin{quote}
\small
Let's carefully go step-by-step.

\textbf{Step 1: Represent coordinates.}

Bill's starting point is $(0,0)$.

\textbf{Step 2: First movement.}

He walks $\frac{1}{2}$ mile south, taking him to
\[
\left(0,-\frac{1}{2}\right).
\]

\textbf{Step 3: Second movement.}

He walks $\frac{3}{4}$ mile east, taking him to
\[
\left(\frac{3}{4},-\frac{1}{2}\right).
\]

\textbf{Step 4: Third movement.}

He walks another $\frac{1}{2}$ mile south:
\[
\left(\frac{3}{4},-\frac{1}{2}-\frac{1}{2}\right)
=
\left(\frac{3}{4},-1\right).
\]

\textbf{Step 5: Distance from the starting point.}

\[
\text{Distance}
=
\sqrt{\left(\frac{3}{4}\right)^2+(-1)^2}
=
\sqrt{\frac{9}{16}+1}
=
\sqrt{\frac{25}{16}}
=
\frac{5}{4}
=
1.25.
\]

\textbf{Final answer:}
\[
\boxed{1.25}
\]
\end{quote}

\newpage

\section{Fixed style Pass@$k$}
\label{app:fixed-style-passk}
\begin{figure}[htbp]
    \centering
    \includegraphics[width=\textwidth]{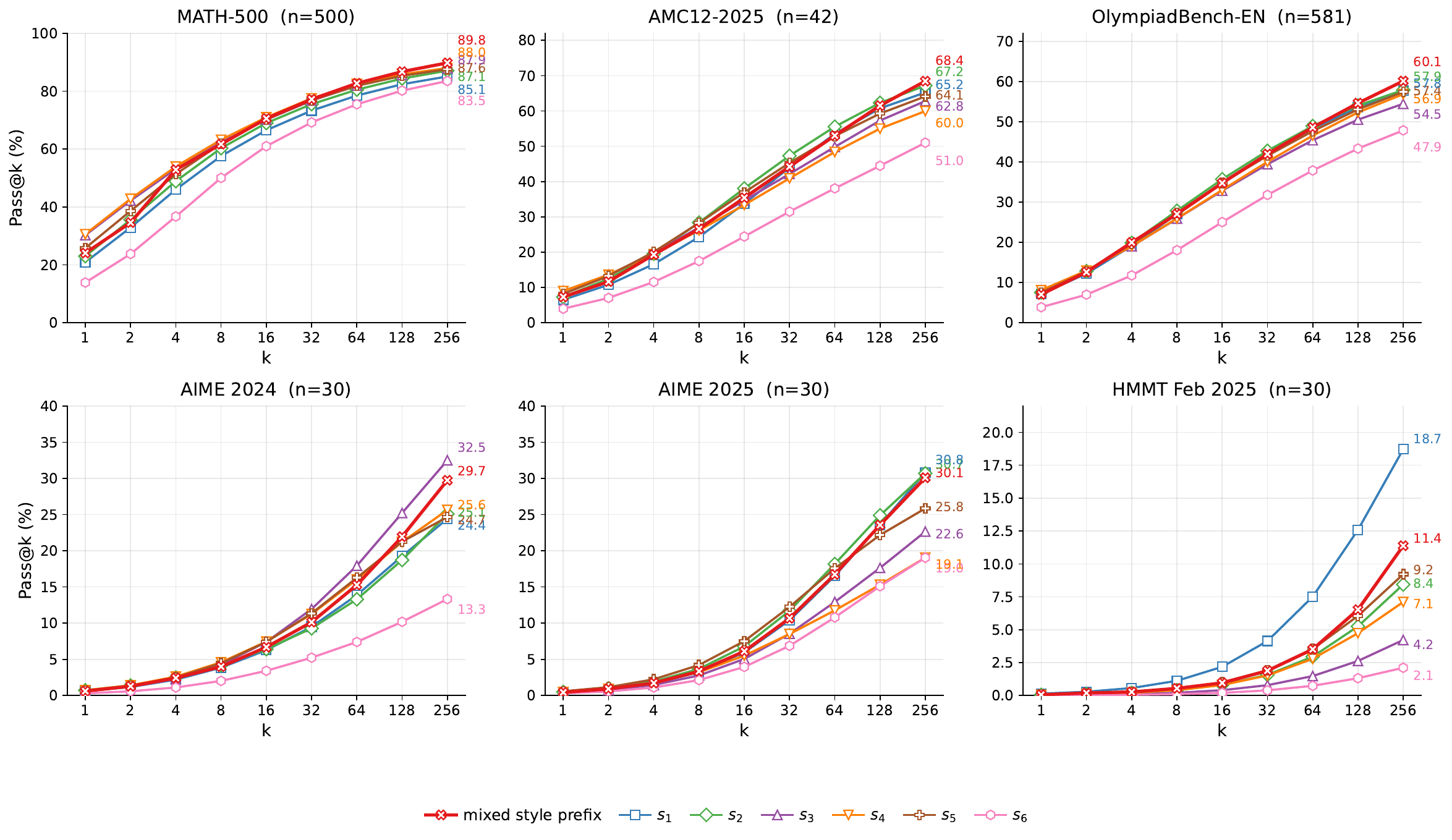}
    \caption{Qwen3 0.6B conditioning individual styles}
    \label{fig:0p6bpassk_style}
\end{figure}
\begin{figure}[htbp]
    \centering
    \includegraphics[width=\textwidth]{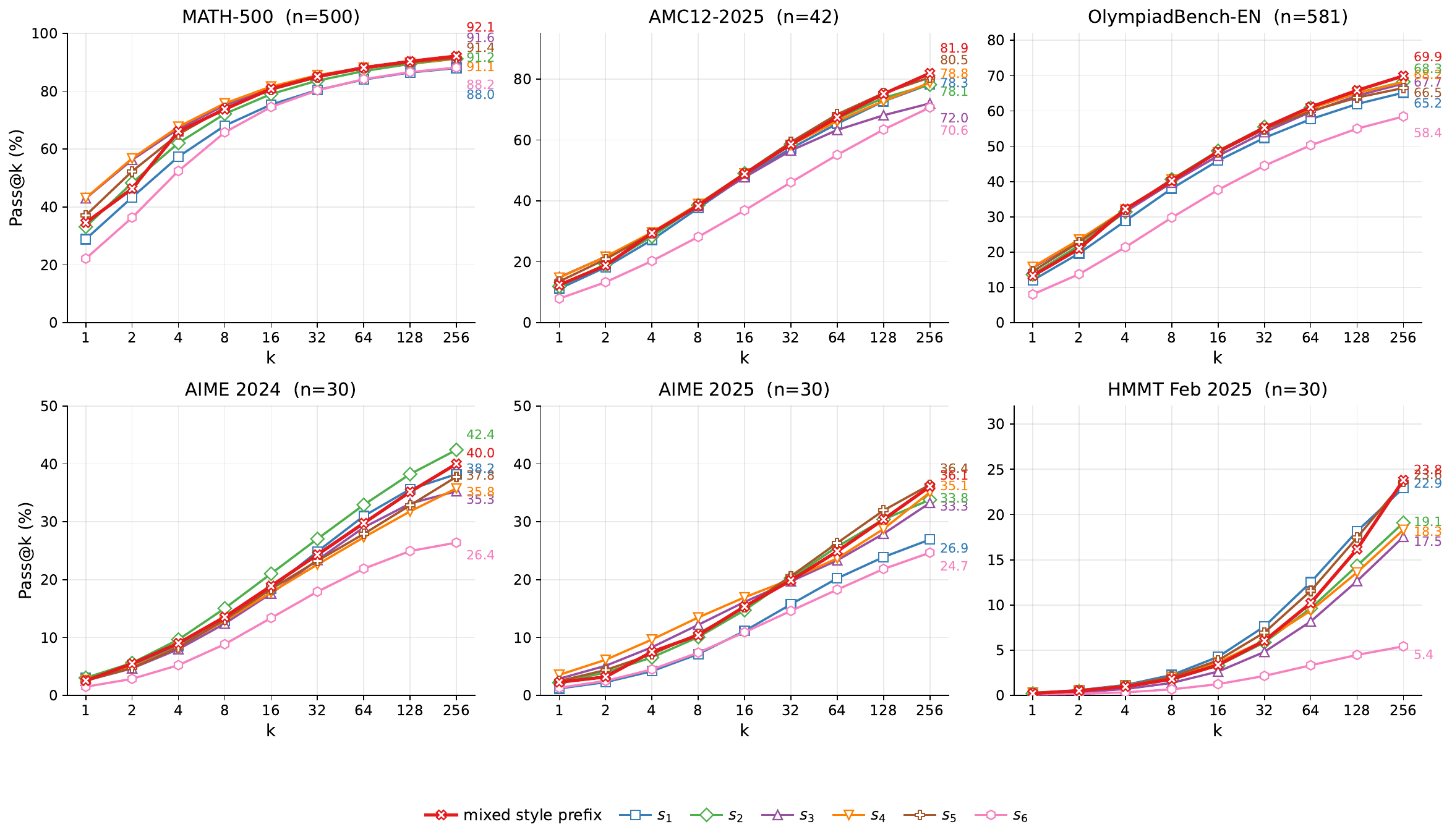}
    \caption{Qwen3 1.7B conditioning individual styles}
    \label{fig:1p7bpassk_style}
\end{figure}
\begin{figure}[htbp]
    \centering
    \includegraphics[width=\textwidth]{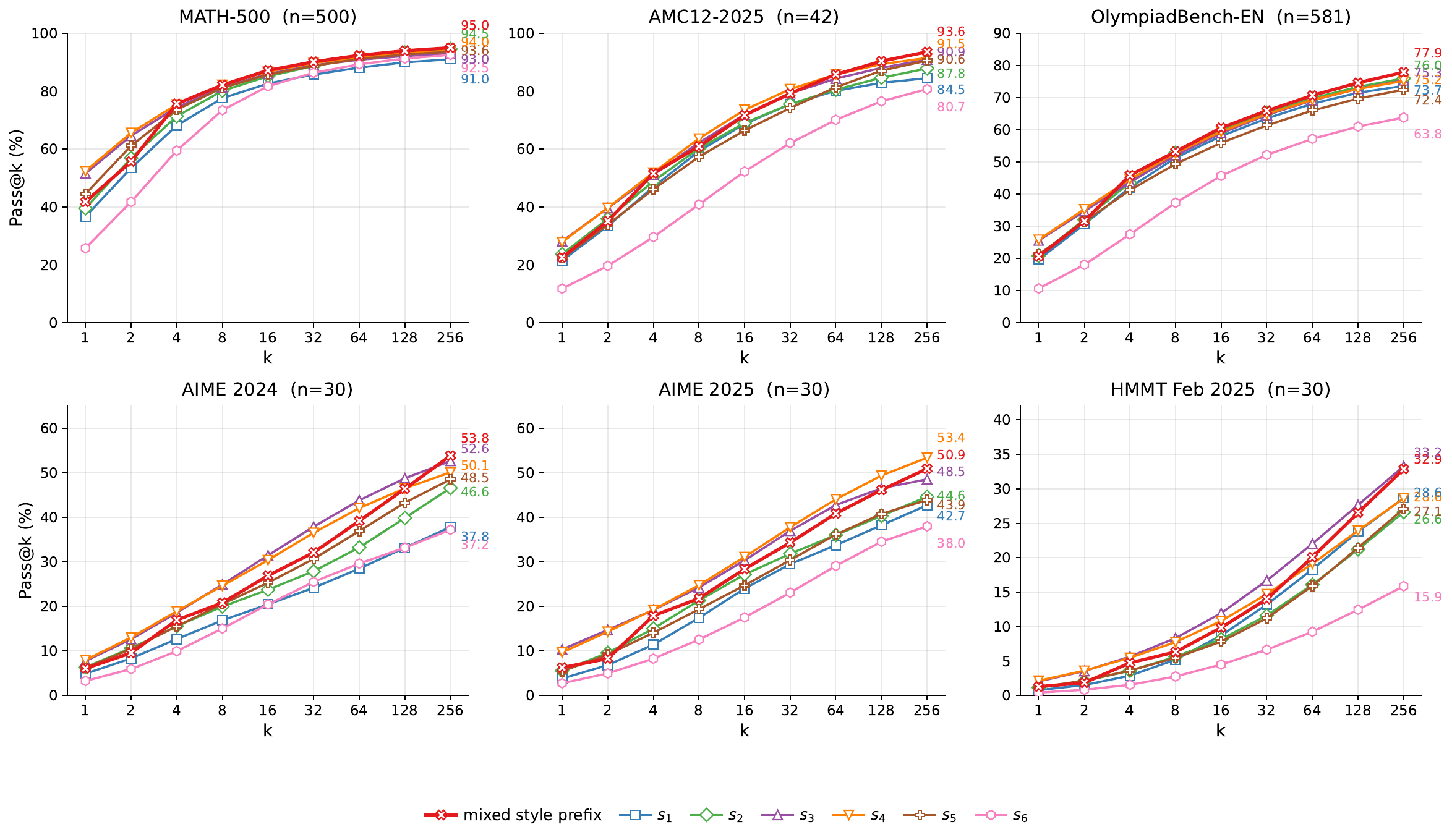}
    \caption{Qwen3 4B conditioning individual styles}
    \label{fig:4bpassk_style}
\end{figure}
\end{document}

%% file: math_commands.tex
\usepackage{amsmath,amsfonts,bm}

\def\eqref#1{equation~\ref{#1}}

\def\1{\bm{1}}

\DeclareMathAlphabet{\mathsfit}{\encodingdefault}{\sfdefault}{m}{sl}
\SetMathAlphabet{\mathsfit}{bold}{\encodingdefault}{\sfdefault}{bx}{n}



%% file: iclr2027_conference.bib
@article{hu2026scas,
  title={The Strongest Teacher Is Not Always the Best Teacher: Student-Centric Answer Selection},
  author={Hu, Zhengyu and Xiao, Zheyuan and Song, Linxin and Jiang, Fengqing and Li, Yuetai and Xiong, Zhihan and Liu, Yue and Lin, Junhao and Su, Yao and Hu, Lijie and Ding, Kaize and Xiao, Teng and Poovendran, Radha},
  journal={arXiv preprint arXiv:2605.26872},
  year={2026}
}

@article{cobbe2021gsm8k,
  title={Training Verifiers to Solve Math Word Problems},
  author={Cobbe, Karl and Kosaraju, Vineet and Bavarian, Mohammad and Chen, Mark and Jun, Heewoo and Kaiser, Lukasz and Plappert, Matthias and Tworek, Jerry and Hilton, Jacob and Nakano, Reiichiro and others},
  journal={arXiv preprint arXiv:2110.14168},
  year={2021}
}

@inproceedings{john2019disentangled,
    title = "Disentangled Representation Learning for Non-Parallel Text Style Transfer",
    author = "John, Vineet  and
      Mou, Lili  and
      Bahuleyan, Hareesh  and
      Vechtomova, Olga",
    editor = "Korhonen, Anna  and
      Traum, David  and
      M{\`a}rquez, Llu{\'i}s",
    booktitle = "Proceedings of the 57th Annual Meeting of the Association for Computational Linguistics",
    month = jul,
    year = "2019",
    address = "Florence, Italy",
    publisher = "Association for Computational Linguistics",
    url = "https://aclanthology.org/P19-1041/",
    doi = "10.18653/v1/P19-1041",
    pages = "424--434",
}

@article{qwen3,
  title={{Qwen3} Technical Report},
  author={Qwen Team},
  journal={arXiv preprint arXiv:2505.09388},
  year={2025}
}

@inproceedings{
wang2023selfconsistency,
title={Self-Consistency Improves Chain of Thought Reasoning in Language Models},
author={Xuezhi Wang and Jason Wei and Dale Schuurmans and Quoc V Le and Ed H. Chi and Sharan Narang and Aakanksha Chowdhery and Denny Zhou},
booktitle={The Eleventh International Conference on Learning Representations },
year={2023},
url={https://openreview.net/forum?id=1PL1NIMMrw}
}

@article{bardes2022vicreg,
  title={{VICReg}: Variance-Invariance-Covariance Regularization for Self-Supervised Learning},
  author={Bardes, Adrien and Ponce, Jean and LeCun, Yann},
  journal={ICLR},
  year={2022}
}

@inproceedings{zbontar2021barlow,
  title={Barlow Twins: Self-Supervised Learning via Redundancy Reduction},
  author={Zbontar, Jure and Jing, Li and Misra, Ishan and LeCun, Yann and Deny, St{\'e}phane},
  booktitle={ICML},
  year={2021}
}

@inproceedings{chen2020simclr,
  title={A Simple Framework for Contrastive Learning of Visual Representations},
  author={Chen, Ting and Kornblith, Simon and Norouzi, Mohammad and Hinton, Geoffrey},
  booktitle={ICML},
  year={2020}
}

@article{hendrycks2021math,
  title={Measuring Mathematical Problem Solving With the {MATH} Dataset},
  author={Hendrycks, Dan and Burns, Collin and Kadavath, Saurav and Arora, Akul and Basart, Steven and Tang, Eric and Song, Dawn and Steinhardt, Jacob},
  journal={NeurIPS Datasets and Benchmarks},
  year={2021}
}

@misc{deepscaler2025,
  title={DeepScaleR: Surpassing {O1}-Preview with a 1.5{B} Model by Scaling {RL}},
  author={Luo, Michael and Tan, Sijun and Wong, Justin and Shi, Xiaoxiang and Tang, William and Roongta, Manan and Cai, Colin and Luo, Jeffrey and Zhang, Tianjun and Li, Erran and Popa, Raluca Ada and Stoica, Ion},
  year={2025},
  howpublished={Notion Blog},
  url={https://pretty-radio-b75.notion.site/DeepScaleR-Surpassing-O1-Preview-with-a-1-5B-Model-by-Scaling-RL-19681902c1468005bed8ca303013a4e2}
}

@misc{openr12025,
  title={Open-{R1}: a fully open reproduction of {DeepSeek}-{R1}},
  author={{Hugging Face}},
  year={2025},
  howpublished={\url{https://github.com/huggingface/open-r1}}
}

@inproceedings{
wu2025modeconditioning,
title={Mode-conditioning unlocks superior test-time compute scaling},
author={Chen Henry Wu and Sachin Goyal and Aditi Raghunathan},
booktitle={The Fourteenth International Conference on Learning Representations},
year={2026},
url={https://openreview.net/forum?id=JzkdJQzPw1}
}

@inproceedings{lippmann2025stylesubstance,
  title={Style over Substance: Distilled Language Models Reason Via Stylistic Replication},
  author={Lippmann, Philip and Yang, Jie},
  booktitle={Proceedings of the Second Conference on Language Modeling},
  year={2025}
}

@inproceedings{shen2017style,
  title={Style Transfer from Non-Parallel Text by Cross-Alignment},
  author={Shen, Tianxiao and Lei, Tao and Barzilay, Regina and Jaakkola, Tommi},
  booktitle={Advances in Neural Information Processing Systems},
  volume={30},
  year={2017}
}

@inproceedings{bachmann2026potential,
  title={The Potential of CoT for Reasoning: A Closer Look at Trace Dynamics},
  author={Bachmann, Gregor and Jiang, Yichen and Moosavi-Dezfooli, Seyed-Mohsen and Nabi, Moin},
  booktitle={International Conference on Learning Representations},
  year={2026}
}

@article{huang2026finetune,
  title={How to Fine-Tune a Reasoning Model? A Teacher--Student Cooperation Framework to Synthesize Student-Consistent SFT Data},
  author={Huang, Zixian and Yang, Kaichen and Huang, Xu and Hao, Feiyang and Ge, Qiming and Li, Bowen and Du, He and Chen, Kai and Guo, Qipeng},
  journal={arXiv preprint arXiv:2604.14164},
  year={2026}
}

@article{chen2025thoughts,
  title={Your Thoughts Tell Who You Are: Characterize the Reasoning Patterns of LRMs},
  author={Chen, Yida and Mao, Yuning and Yang, Xianjun and Ge, Suyu and Bi, Shengjie and Liu, Lijuan and Hosseini, Saghar and Tan, Liang and Nie, Yixin and Nie, Shaoliang},
  journal={arXiv preprint arXiv:2509.24147},
  year={2025}
}

@inproceedings{zhang2026overthinking,
  title={Do LLMs Really Need 10+ Thoughts for ``Find the Time 1000 Days Later''? Towards Structural Understanding of LLM Overthinking},
  author={Zhang, Xinliang Frederick and Mohananey, Anhad and Chronopoulou, Alexandra and Papalampidi, Pinelopi and Gupta, Somit and Munkhdalai, Tsendsuren and Wang, Lu and Upadhyay, Shyam},
  booktitle={Proceedings of the 64th Annual Meeting of the Association for Computational Linguistics},
  year={2026},
  pages={17005--17030},
  publisher={Association for Computational Linguistics},
  doi={10.18653/v1/2026.acl-long.773}
}

@inproceedings{zhu2024styleflow,
  title={StyleFlow: Disentangle Latent Representations via Normalizing Flow for Unsupervised Text Style Transfer},
  author={Zhu, Kangchen and Tian, Zhiliang and Wei, Jingyu and Luo, Ruifeng and Song, Yiping and Mao, Xiaoguang},
  booktitle={Proceedings of the 2024 Joint International Conference on Computational Linguistics, Language Resources and Evaluation},
  pages={15384--15397},
  year={2024}
}

@article{oord2018cpc,
  title={Representation Learning with Contrastive Predictive Coding},
  author={van den Oord, Aaron and Li, Yazhe and Vinyals, Oriol},
  journal={arXiv preprint arXiv:1807.03748},
  year={2018}
}

@inproceedings{cheng2020information,
  title={Improving Disentangled Text Representation Learning with Information-Theoretic Guidance},
  author={Cheng, Pengyu and Min, Martin Renqiang and Shen, Dinghan and Malon, Christopher and Zhang, Yizhe and Li, Yitong and Carin, Lawrence},
  booktitle={Proceedings of the 58th Annual Meeting of the Association for Computational Linguistics},
  pages={7530--7541},
  year={2020},
  publisher={Association for Computational Linguistics},
  doi={10.18653/v1/2020.acl-main.673}
}

@article{sha2024closedloop,
    title = "Text Attribute Control via Closed-Loop Disentanglement",
    author = "Sha, Lei  and
      Lukasiewicz, Thomas",
    journal = "Transactions of the Association for Computational Linguistics",
    volume = "12",
    year = "2024",
    address = "Cambridge, MA",
    publisher = "MIT Press",
    url = "https://aclanthology.org/2024.tacl-1.11/",
    doi = "10.1162/tacl_a_00640",
    pages = "190--209",
}

@inproceedings{han2024disentangled,
  title={Disentangled Learning with Synthetic Parallel Data for Text Style Transfer},
  author={Han, Jingxuan and Wang, Quan and Guo, Zikang and Xu, Benfeng and Zhang, Licheng and Mao, Zhendong},
  booktitle={Proceedings of the 62nd Annual Meeting of the Association for Computational Linguistics},
  pages={15187--15201},
  year={2024},
  publisher={Association for Computational Linguistics},
  doi={10.18653/v1/2024.acl-long.811}
}

@inproceedings{hsieh2023distilling,
  title={Distilling Step-by-Step! Outperforming Larger Language Models with Less Training Data and Smaller Model Sizes},
  author={Hsieh, Cheng-Yu and Li, Chun-Liang and Yeh, Chih-kuan and Nakhost, Hootan and Fujii, Yasuhisa and Ratner, Alex and Krishna, Ranjay and Lee, Chen-Yu and Pfister, Tomas},
  booktitle={Findings of the Association for Computational Linguistics: ACL 2023},
  pages={8003--8017},
  year={2023},
  publisher={Association for Computational Linguistics},
  doi={10.18653/v1/2023.findings-acl.507}
}

@article{mukherjee2023orca,
  title={Orca: Progressive Learning from Complex Explanation Traces of GPT-4},
  author={Mukherjee, Subhabrata and Mitra, Arindam and Jawahar, Ganesh and Agarwal, Sahaj and Palangi, Hamid and Awadallah, Ahmed},
  journal={arXiv preprint arXiv:2306.02707},
  year={2023}
}

@InProceedings{feng2024kpod,
  title = 	 {Keypoint-based Progressive Chain-of-Thought Distillation for {LLM}s},
  author =       {Feng, Kaituo and Li, Changsheng and Zhang, Xiaolu and Zhou, Jun and Yuan, Ye and Wang, Guoren},
  booktitle = 	 {Proceedings of the 41st International Conference on Machine Learning},
  pages = 	 {13241--13255},
  year = 	 {2024},
  editor = 	 {Salakhutdinov, Ruslan and Kolter, Zico and Heller, Katherine and Weller, Adrian and Oliver, Nuria and Scarlett, Jonathan and Berkenkamp, Felix},
  volume = 	 {235},
  series = 	 {Proceedings of Machine Learning Research},
  month = 	 {21--27 Jul},
  publisher =    {PMLR},
  url = 	 {https://proceedings.mlr.press/v235/feng24e.html},
}

@inproceedings{allal2025smollm2smolgoesbig,
      title={SmolLM2: When Smol Goes Big -- Data-Centric Training of a Small Language Model}, 
      author={Loubna Ben Allal and Anton Lozhkov and Elie Bakouch and Gabriel Martín Blázquez and Guilherme Penedo and Lewis Tunstall and Andrés Marafioti and Hynek Kydlíček and Agustín Piqueres Lajarín and Vaibhav Srivastav and Joshua Lochner and Caleb Fahlgren and Xuan-Son Nguyen and Clémentine Fourrier and Ben Burtenshaw and Hugo Larcher and Haojun Zhao and Cyril Zakka and Mathieu Morlon and Colin Raffel and Leandro von Werra and Thomas Wolf},
      year={2025},
      booktitle={Second Conference on Language Modeling}
}

@inproceedings{
jia2026training,
title={Training Large Language Models To Reason In Parallel With Global Forking Tokens},
author={Sheng Jia and Xiao Wang and Shiva Kasiviswanathan},
booktitle={The Fourteenth International Conference on Learning Representations},
year={2026},
url={https://openreview.net/forum?id=xBQvvkg4Wc}
}

@misc{makino2025supervisedcontrastiveblockdisentanglement,
      title={Supervised Contrastive Block Disentanglement}, 
      author={Taro Makino and Ji Won Park and Natasa Tagasovska and Takamasa Kudo and Paula Coelho and Jan-Christian Huetter and Heming Yao and Burkhard Hoeckendorf and Ana Carolina Leote and Stephen Ra and David Richmond and Kyunghyun Cho and Aviv Regev and Romain Lopez},
      year={2025},
      eprint={2502.07281},
      archivePrefix={arXiv},
      primaryClass={cs.LG},
      url={https://arxiv.org/abs/2502.07281}, 
}

@misc{gemini31pro2026,
  title        = {Gemini 3.1 Pro Model Card},
  author       = {{Google DeepMind}},
  year         = {2026},
  month        = {February},
  url          = {https://deepmind.google/models/model-cards/gemini-3-1-pro/},
  note         = {Accessed: 2026-09-25}
}

@misc{grok46modelcard2026,
  title        = {Grok 4.6 Model Card},
  author       = {{xAI}},
  year         = {2026},
  month        = {August},
  url          = {https://media.x.ai/v1/website/card-4p6-4cd2dc57.pdf},
  note         = {Accessed: 2026-09-25}
}

@article{gemma4techreport,
  title        = {Gemma 4 Technical Report},
  author       = {{Gemma Team Google DeepMind}},
  year         = {2026},
  journal      = {arXiv preprint arXiv:2607.02770},
  eprint       = {2607.02770},
  archivePrefix= {arXiv},
  primaryClass = {cs.CL}
}

@techreport{openai2025gpt5,
  title       = {GPT-5 System Card},
  author      = {{OpenAI}},
  institution = {OpenAI},
  year        = {2025},
  month       = {August},
  url         = {https://cdn.openai.com/gpt-5-system-card.pdf}
}

@article{openai2025gptoss,
  title        = {gpt-oss-120b \& gpt-oss-20b Model Card},
  author       = {{OpenAI}},
  year         = {2025},
  journal      = {arXiv preprint arXiv:2508.10925},
  eprint       = {2508.10925},
  archivePrefix= {arXiv},
  primaryClass = {cs.CL}
}

@article{dubey2024llama3,
  title        = {The Llama 3 Herd of Models},
  author       = {Dubey, Abhimanyu and Jauhri, Abhinav and Pandey, Abhinav and Kadian, Abhishek and Al-Dahle, Ahmad and Letman, Aiesha and Mathur, Akhil and Schelten, Alan and Yang, Amy and Fan, Angela and others},
  year         = {2024},
  journal      = {arXiv preprint arXiv:2407.21783},
  eprint       = {2407.21783},
  archivePrefix= {arXiv},
  primaryClass = {cs.AI}
}

@article{teamolmo2025olmo3,
  title        = {Olmo 3},
  author       = {Ettinger, Allyson and {Team Olmo} and {Allen Institute for AI}},
  year         = {2025},
  journal      = {arXiv preprint arXiv:2512.13961},
  eprint       = {2512.13961},
  archivePrefix= {arXiv},
  primaryClass = {cs.CL}
}

@article{abdin2025phi4reasoning,
  title        = {Phi-4-reasoning Technical Report},
  author       = {Abdin, Marah and Aneja, Jyoti and Awadalla, Hany and Awadallah, Ahmed and Bach, Nguyen and Bahree, Amit and Behl, Arushi and Benhaim, Alon and Bhalerao, Rahul and Bubeck, S{\'e}bastien and others},
  year         = {2025},
  journal      = {arXiv preprint arXiv:2504.21318},
  eprint       = {2504.21318},
  archivePrefix= {arXiv},
  primaryClass = {cs.AI}
}

@article{yang2024qwen25,
  title        = {Qwen2.5 Technical Report},
  author       = {Yang, An and Yang, Baosong and Zhang, Beichen and Hui, Binyuan and Zheng, Bo and Yu, Bo-Wen and Li, Chengyuan and Liu, Dayiheng and Huang, Fei and Dong, Guanting and others},
  year         = {2024},
  journal      = {arXiv preprint arXiv:2412.15115},
  eprint       = {2412.15115},
  archivePrefix= {arXiv},
  primaryClass = {cs.CL}
}

@article{qwenteam2025qwen3,
  title        = {Qwen3 Technical Report},
  author       = {{Qwen Team}},
  year         = {2025},
  journal      = {arXiv preprint arXiv:2505.09388},
  eprint       = {2505.09388},
  archivePrefix= {arXiv},
  primaryClass = {cs.CL}
}

@article{qwenteam2026qwen35,
  title        = {Qwen3.5-Omni Technical Report},
  author       = {{Qwen Team}},
  year         = {2026},
  journal      = {arXiv preprint arXiv:2604.15804},
  eprint       = {2604.15804},
  archivePrefix= {arXiv},
  primaryClass = {cs.CL}
}
